\documentclass[pdflatex,sn-nature]{sn-jnl}

\usepackage{graphicx}
\usepackage{multirow}
\usepackage{amsmath,amssymb,amsfonts}
\usepackage{amsthm}
\usepackage[title]{appendix}
\usepackage{xcolor}
\usepackage{textcomp}
\usepackage{manyfoot}
\usepackage{booktabs}
\usepackage{algorithmicx}
\usepackage{algpseudocode}
\usepackage{listings}
\usepackage{etoolbox}

\makeatletter
\patchcmd{\@maketitle}
  {Corresponding author(s). E-mail(s):}
  {Correspondence:}
  {}
  {\PackageWarning{sn-jnl-custom}{Could not patch the correspondence label}}
\makeatother

\AtBeginDocument{%
  \hypersetup{
    pdftitle={Reproducible macroscopic dynamics in a closed-loop human-AI learning system},
    pdfauthor={Minlin Wu, Xu Fang, Yicheng Zhang, Chenyu Zhou, Zhiyi Liu},
    pdfsubject={Macroscopic dynamics in closed-loop human-AI learning systems},
    pdfkeywords={closed-loop human-AI systems, macroscopic dynamics, mechanism model, self-supervised learning}
  }%
}

\begin{document}

\title{Reproducible macroscopic dynamics in a closed-loop human--AI learning system}   

\author*[1]{\fnm{Minlin} \sur{Wu}}
\email{evanwuminlin@gmail.com}

\author[1]{\fnm{Xu} \sur{Fang}}

\author[2]{\fnm{Yicheng} \sur{Zhang}}

\author[1]{\fnm{Chenyu} \sur{Zhou}}

\author*[1]{\fnm{Zhiyi} \sur{Liu}}
\email{zhiyil696@gmail.com}

\affil*[1]{%
  \orgdiv{Tianli Qiming AI Research Institute},
  \orgname{Sichuan Qiming Daren Technology Co., Ltd.},
  \orgaddress{\city{Chengdu}, \country{China}}%
}

\affil[2]{%
  \orgname{Swiss AI Laboratories},
  \orgaddress{\city{Blonay}, \country{Switzerland}}%
}

\abstract{
Closed-loop human--AI systems generate high-dimensional behavioural trajectories whose collective dynamics remain obscure. Using 297,915 learners' adaptive-tutoring histories, we define semantic order variables before model fitting and test them in user-disjoint cohorts. The state exhibits reproducible basin-like flow and operationally defined, state-heterogeneous metastable-like kinetics. A construction-matched null distinguishes normalised-memory relaxation from a reproducible excess field. A four-term conditional mechanism recovers population drift (\(r=0.946\); learner-bootstrap 95\% CI, \(0.935\)--\(0.955\)). Predictive event-level self-supervised learning recovers the state and learned-plane flow; null-referenced corrections retain directional, partial-amplitude excess-field structure without full calibration. Shuffled-order training reverses learned-plane flow on ordered trajectories; support-alignment randomisation selectively reduces inward transport. Both axes remain linearly accessible without state supervision. Without cross-model fitting, the models share leading population drift (\(r=0.866\); learner-bootstrap 95\% CI, \(0.857\)--\(0.875\)) and persistence ordering; residual directions remain model-specific. These results identify an externally anchored leading-order effective field linking empirical dynamics, an interpretable mechanism and neural computation.
}

\maketitle

Modern artificial intelligence systems increasingly operate within closed feedback loops, in which users, tasks, feedback and platform content jointly shape future observations. The resulting records are trajectories of partially observed human--algorithm systems rather than static samples from an external distribution \cite{Perdomo2020,Glickman2025}. This setting raises a question beyond next-event prediction: do such event streams admit reproducible macroscopic variables that support an effective dynamical description?

This question reflects a broader problem in complex systems: identifying collective variables that emerge from many microscopic degrees of freedom yet have reproducible dynamics of their own. High-dimensional observations can admit low-dimensional coordinates that expose invariant structure, stability or dominant modes \cite{Barnett2023,Moore2025}. Recent studies have learned thermodynamic coordinates and task-oriented reduced models from complex data, while predictive learning can expose latent variables in low-dimensional neural representations \cite{Chen2024,Fabiani2024,Recanatesi2021}. For closed-loop behavioural data, the relevant test is whether observable event semantics define macrostates with reproducible held-out dynamics, low-order closure and accessibility in predictive event-level representations \cite{Lu2025}.

Closed-loop behavioural logs do not arise from controlled physical experiments: learners interact with content, feedback and platform policies without a known microscopic equation of motion. We therefore seek semantic coordinates that organise heterogeneous events into reproducible occupancy, flow and kinetics while separating empirical structure from relaxation implied by their normalised-memory construction. These coordinates provide a common macroscopic level for empirical, mechanistic and neural analyses.

\begin{figure*}
    \centering
    \includegraphics[width=\textwidth]{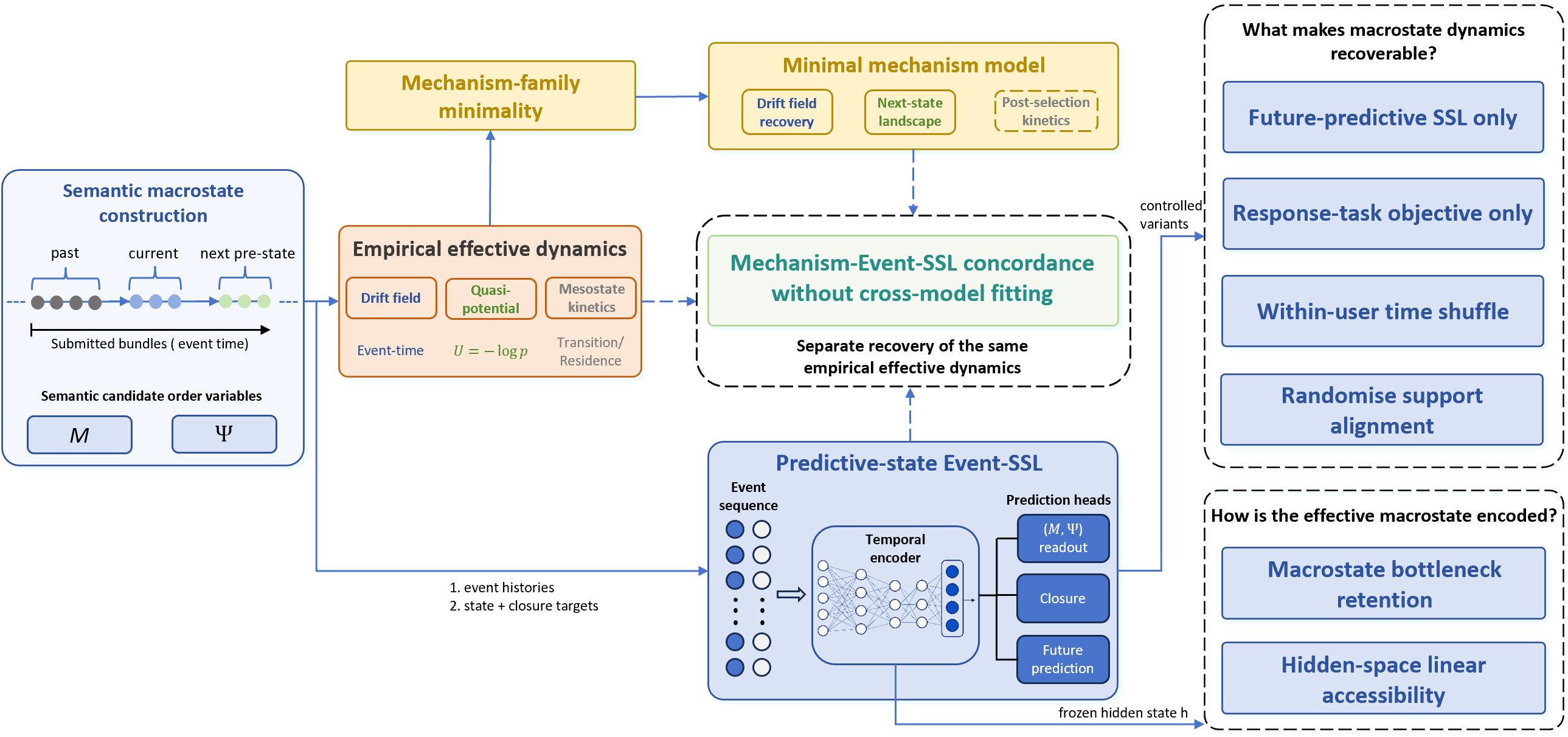}
    \caption{\textbf{Schematic overview.} Submitted-bundle histories define \(\mathbf X_n=(M_n,\Psi_n)\), with response order \(M\) and exposure alignment \(\Psi\); the next bundle defines the one-step transition. The empirical branch estimates user-balanced occupancy, an occupancy-derived quasi-potential, event-time drift and mesostate kinetics. A prespecified hierarchy selects a conditional one-step mechanism without using transition or residence diagnostics. Event-SSL combines future prediction with macrostate and closure objectives. Temporal-order and support-alignment controls test dynamical validity; hidden-state analyses assess bottleneck retention and linear accessibility. The mechanism and Event-SSL are compared with the same empirical dynamics without cross-model fitting.}
    \label{fig:schematic_view}
\end{figure*}

Adaptive learning platforms provide a natural testbed. EdNet comprises 131,441,538 interactions from 784,309 learners at four behavioural resolutions; its most detailed level, EdNet-KT4, records question-solving, revision, study, multimedia and access events for the 297,915 learners analysed here \cite{Choi2020}. EdNet was introduced for knowledge tracing and learning-path recommendation, but its heterogeneous actions also allow us to test whether learner--platform trajectories organise around reproducible macroscopic states.

We construct candidate order variables by treating event semantics as measurement constraints rather than generic features. Responses provide signed evidence relative to item-conditioned expectations; question-solving, study and support events quantify exposure and its alignment with unresolved demand, while access and process events remain contextual. This preserves distinct response and exposure signals rather than collapsing heterogeneous actions into a single prediction target.

We evaluate this construction in three stages (Fig.~\ref{fig:schematic_view}). First, we test whether occupancy, flow and coarse kinetics reproduce across user-disjoint cohorts. Second, we select and freeze a compact conditional mechanism within a prespecified family hierarchy, reserving transition and residence for post-selection tests. Third, predictive event-level self-supervised learning (Event-SSL) combines future prediction with auxiliary macrostate and one-step closure objectives. Temporal-order and support-alignment perturbations then test whether event histories support a state that is not only decodable but dynamically self-consistent in the model's own coordinates.

The analyses support a reproducible effective field beyond construction-implied relaxation. The family-bounded mechanism provides calibrated conditional closure, whereas Event-SSL recovers self-consistent learned-plane dynamics and a linearly accessible state \cite{Nanda2023,Park2024}. Without cross-model fitting, the models agree on leading population drift and persistence ordering while retaining model-specific residual structure. This provides a closed-loop behavioural instance of effective macroscopic dynamics \cite{Simon2026}.

\section*{Results}

\begin{figure*}
    \centering
    \includegraphics[width=\textwidth]{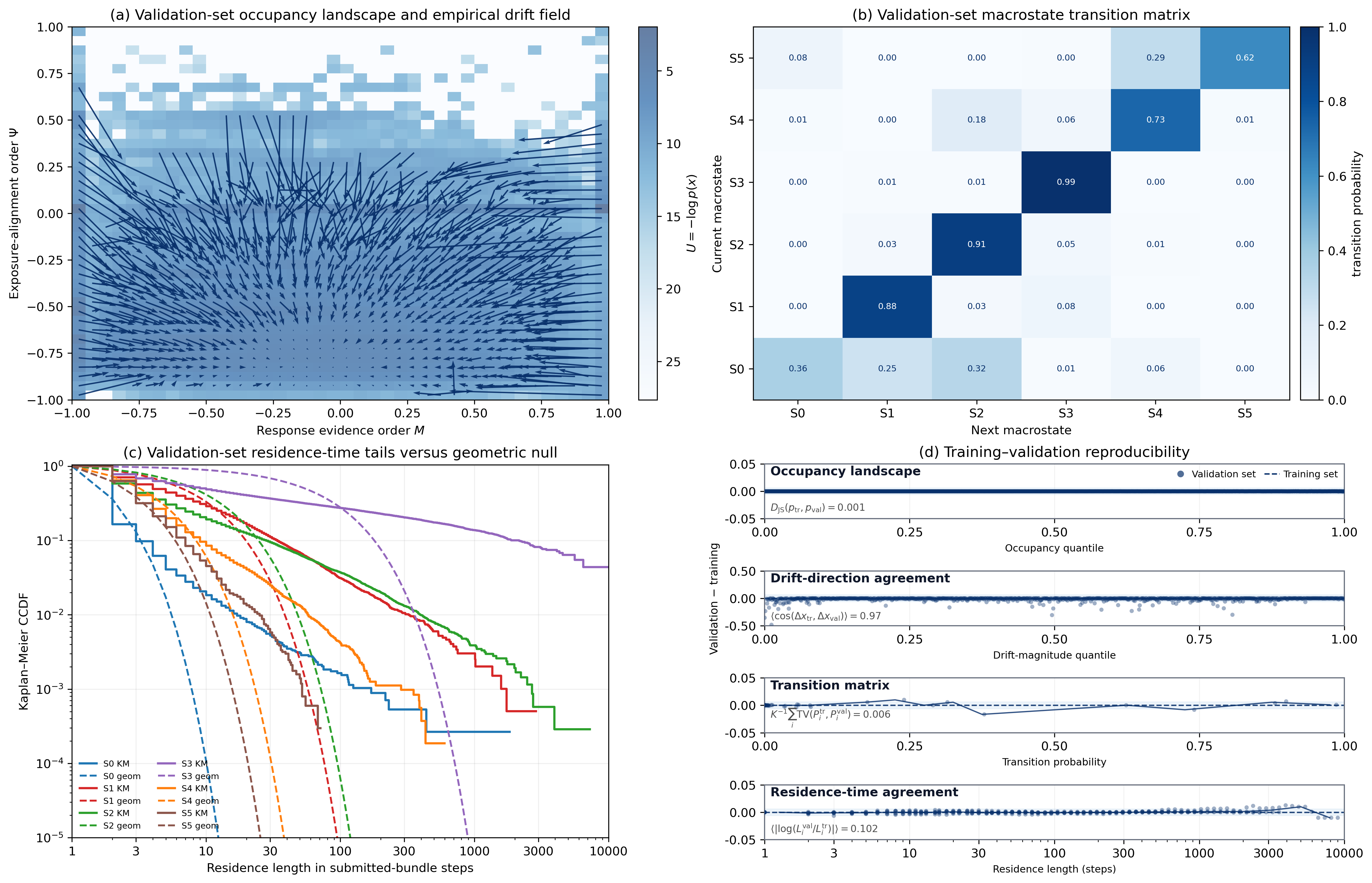}
    \caption{\textbf{Empirical effective dynamics and operational mesostate kinetics.} \textbf{a}, Validation occupancy-derived quasi-potential \(U=-\log p_{\mathrm{ub}}(M,\Psi)\), overlaid with conditional mean one-step drift; values are clipped at the 98th percentile. \textbf{b}, Row-normalised transitions under the fixed six-mesostate partition. \textbf{c}, Kaplan--Meier residence-time survival curves (solid) and state-matched geometric references \(p_{ii}^{\,\ell-1}\) (dashed) on logarithmic axes. \textbf{d}, Validation-minus-training differences in occupancy, local drift direction, transitions and censor-aware residence tails. Points show matched bins, matrix entries or residence lengths; solid curves are binned medians and dashed lines the training reference. Annotations report occupancy Jensen--Shannon (JS) divergence, mean local drift cosine, mean row-wise total variation (TV) and mean absolute log difference in statewise 10-step tail ratios. Lower values indicate closer agreement except for drift cosine.}
    \label{fig:empirical_effective_dynamics}
\end{figure*}

Semantic decomposition of EdNet-KT4 yielded two bounded order variables prespecified before fitting. Response order \(M\) accumulates signed residual evidence relative to item-conditioned baselines, whereas exposure alignment \(\Psi\) measures whether question-solving, study and support address unresolved demand. Positive and negative \(M\) denote above- and below-baseline evidence. Content tags link response evidence to subsequent exposure; maturity, access and process variables enter only accounting or diagnostic analyses.

Submitted-bundle pre-state trajectories were reconstructed for all 297,915 users and split into user-disjoint training (\(n=178{,}749\)), validation (\(n=59{,}583\)) and confirmation (\(n=59{,}583\)) cohorts. Training supplied item-conditioned baselines and field criteria; population fields used full-panel learner-normalised weights before eligibility filtering. Validation tested empirical reproducibility; confirmation was not accessed until the mechanism and representation specifications were frozen.

\begin{figure*}
    \centering
    \includegraphics[width=\textwidth]{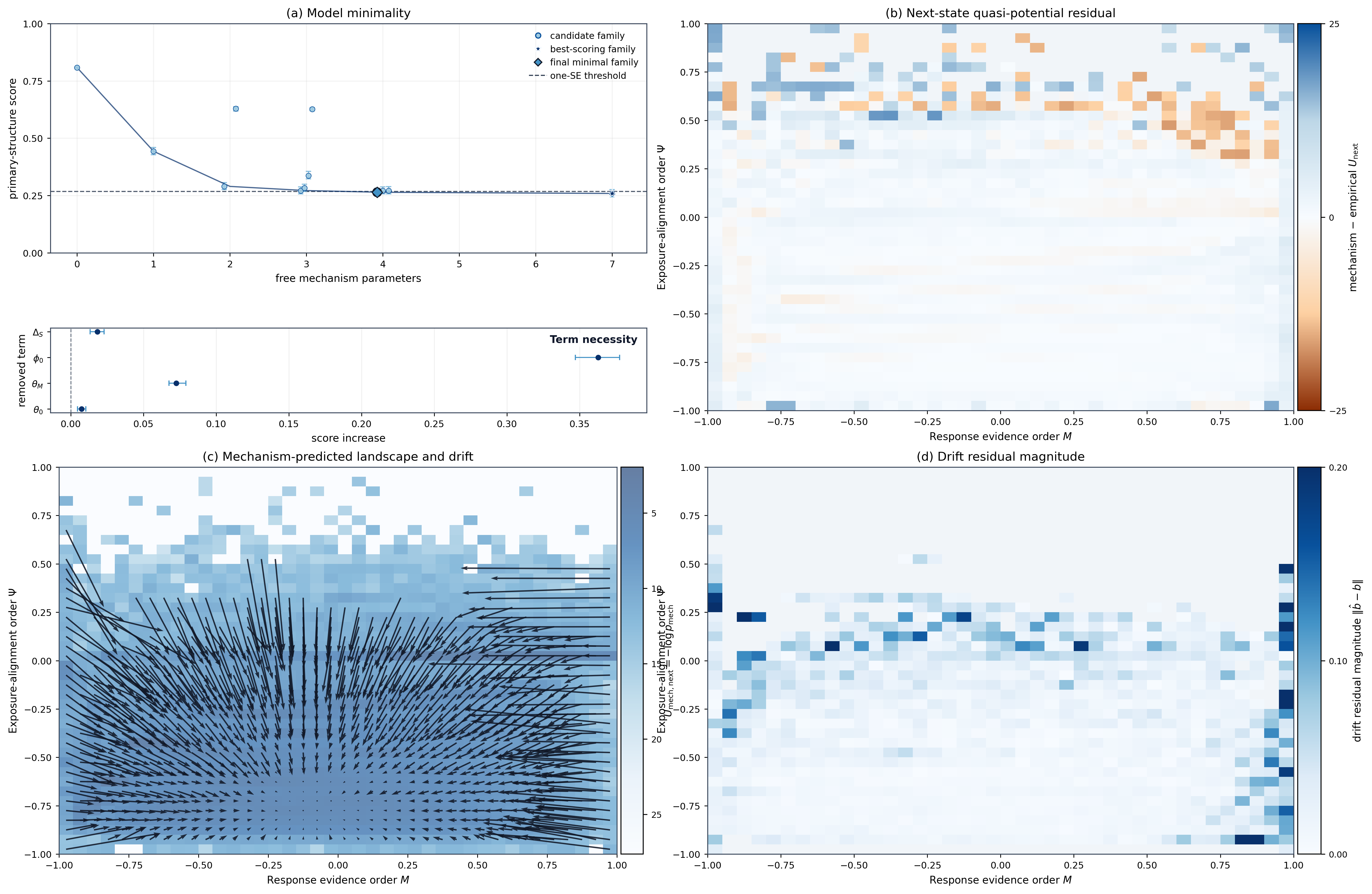}
    \caption{\textbf{Mechanism-family selection and field recovery.} \textbf{a}, Primary-structure score versus substantive parameter count; lower values indicate better recovery. Points and bars show means and 95\% intervals from 300 paired validation-user bootstraps. A star, open circle and diamond mark the best-scoring, selected and frozen families; the dashed line is the one-standard-error threshold. The lower panel shows score changes after refitted one-term deletions under the same eligibility rule. \textbf{b}, Mechanism-minus-empirical next-state quasi-potential, \(\Delta U_{\mathrm{next}}\). \textbf{c}, Mechanism-implied next-state quasi-potential and mean one-step drift. \textbf{d}, Local drift-residual magnitude, \(\|\hat{\mathbf b}-\mathbf b\|\); pale cells lack common support. Panels \textbf{b--d} use 3,233,208 confirmation intervals from 56,195 users.}
    \label{fig:minimal_mechanism_field}
\end{figure*}

\subsection*{Reproducible empirical effective dynamics}

The validation field was predominantly contractive over its supported interior: negative divergence covered 77.6\% of interior user-balanced occupancy, with a weighted mean of \(-0.215\). The field contained 940 supported drift cells; divergence was restricted to complete interior stencils because 27.9\% of occupancy lay in the outermost \(M\) bins. Separately, the occupancy-derived validation quasi-potential had a dominant low-\(\Psi\) minimum, with half the user-balanced mass in \(\Psi\in[-0.875,-0.525]\) (Fig.~\ref{fig:empirical_effective_dynamics}a).

Independently of occupancy, the training field defined a convergence core centred at \((M,\Psi)=(-0.020,-0.582)\). Applied unchanged, it contained 18.5\% of validation user-balanced occupancy; 66.9\% of shell flow was inward and the core-to-shell speed ratio was 0.523. Relative to the construction-matched accounting null, validation supported the full-field departure (Monte Carlo \(p_{\mathrm{MC}}=0.0099\)) and excess negative-divergence occupancy (Benjamini--Hochberg-adjusted \(q=0.0297\)); both replicated in confirmation. Shell inward flow survived adjustment only in confirmation (validation \(q=0.0594\); confirmation \(q=0.0149\)), whereas core slowing remained null-compatible in both cohorts. This localises the replicated excess to field-wide directional structure rather than uniform core slowing. Null-subtracted validation and confirmation fields agreed across 917 cells (vector \(r=0.716\); occupancy-weighted local cosine \(=0.960\); Supplementary Table~1).

In a post hoc denominator-matched comparison with an activity--idle coordinate \((M,\Phi)\), both states reproduced raw fields and departed from their own construction nulls. Relative to this single comparator, the alignment-based state showed approximately threefold greater field-RMS-normalised departure in validation (\(0.3808\) versus \(0.1300\)), a contrast reproduced in confirmation (\(0.3785\) versus \(0.1236\); Supplementary Note~1 and Supplementary Table~2, panel~\textbf{c}).

A \(K=6\) partition fitted on training coordinates and fixed before validation defined mesostates \(S_0\)--\(S_5\), ordered by \(M\) and then \(\Psi\) (Fig.~\ref{fig:empirical_effective_dynamics}b; Supplementary Table~3). Across \(3{,}328{,}409\) validation transitions, all six rows were diagonal-dominant (mean self-transition probability \(0.750\)). Strict user-equal weighting preserved diagonal dominance in five mesostates; only \(S_0\) changed, with self-transition falling from \(0.3623\) to \(0.0860\). This identifies a mesostate-specific weighting boundary rather than a global loss of persistence (Supplementary Note~6 and Supplementary Table~10).

Population-level residence-time distributions were compared with state-matched geometric references using censor-aware Kaplan--Meier estimates (Fig.~\ref{fig:empirical_effective_dynamics}c) \cite{KaplanMeier1958}. The validation cohort contained \(256{,}009\) residence episodes, of which 22.5\% were right-censored. Across state-specific reliable horizons of \(35\)--\(4{,}302\) steps, restricted mean residence time (RMST) exceeded the corresponding geometric reference in all six mesostates by factors of 1.127--6.473 (Supplementary Table~3).

At the prespecified 10-step horizon, Greenwood-variance inference identified positive survival excess in \(S_0\), \(S_4\) and \(S_5\) after Benjamini--Hochberg adjustment (\(q<0.05\)) \cite{Greenwood1926,Benjamini1995}; the same three mesostates remained positive under learner-cluster bootstrap inference, while all six reliable-horizon RMST lift lower bounds exceeded one. The post hoc all-state statistic (mean log RMST lift through 10 steps) lay below the recursive-surrogate 95\% range (\(-0.2280\) versus \(-0.1535\) to \(-0.1489\); upper-tail \(p_{\mathrm{MC}}=1.0000\)), providing no support for a uniform positive 10-step shift. A complementary post hoc maximum-\(t\) (maxT) test identified positive 10-step survival-probability excesses of \(0.0183\)--\(0.0311\) in the same three mesostates after family-wise error control (\(p_{\mathrm{MC}}=0.0099\)). Together, these endpoints separate broad reliable-horizon persistence from positive 10-step excess confined to \(S_0\), \(S_4\) and \(S_5\). RMST lift remained above one in every mesostate for \(K=4\)--8 (Supplementary Note~2 and Supplementary Table~4).

Training and validation dynamics agreed closely (Fig.~\ref{fig:empirical_effective_dynamics}d): occupancy Jensen--Shannon (JS) divergence was \(5.24\times10^{-4}\), while mean local drift cosine was \(0.966\) and drift-speed correlation \(r=0.930\) across 939 common cells \cite{Lin1991}. Mean transition row-wise total variation (TV) was \(0.006\); statewise residence lift and 10-step tail ratios had mean absolute log differences of \(0.057\) and \(0.102\). Binary thresholding affected the exact core contour (Jaccard \(0.438\)), while all 1,000 learner reweightings recovered dynamically qualified regions in both splits and their selection-frequency maps overlapped by 0.785 (Supplementary Note~1). The recurring contractive region was therefore more reproducible than any single thresholded contour.

Replicated basin geometry, learner-cluster-robust RMST lift and construction-aware 10-step survival excess in \(S_0\), \(S_4\) and \(S_5\) support population-level, operationally defined, state-heterogeneous metastable-like kinetics in submitted-bundle event time under the stated Kaplan--Meier censoring assumption. The frozen phase plane served as the common target for mechanism and representation analyses.

\begin{figure*}
    \centering
    \includegraphics[width=\textwidth]{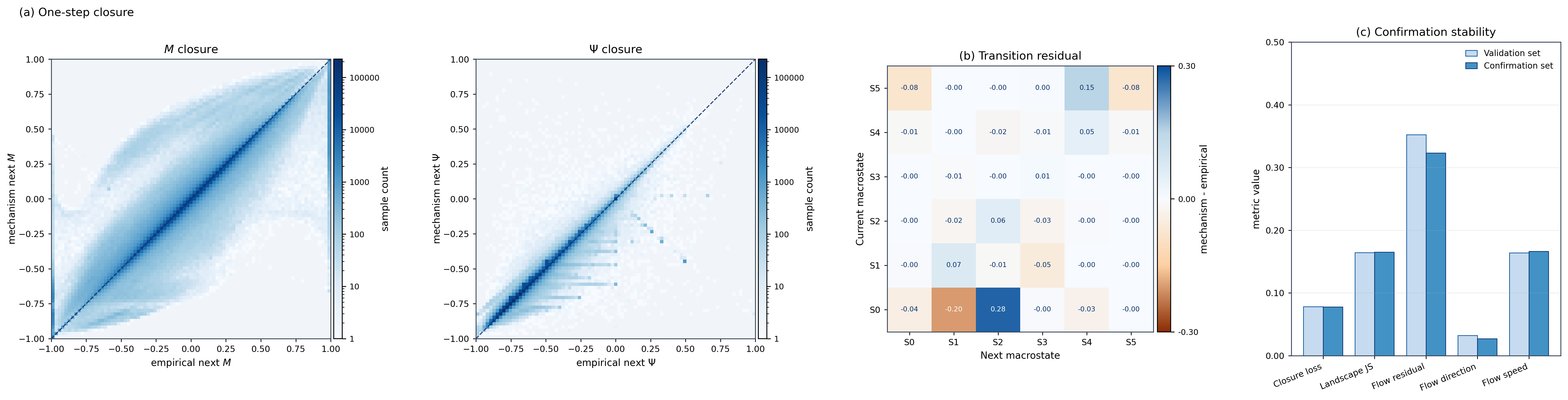}
    \caption{\textbf{One-step closure and post-selection kinetics.} \textbf{a}, Binned empirical versus mechanism-implied next-state \(M\) and \(\Psi\); dashed lines show identity and colour denotes interval count on a logarithmic scale. \textbf{b}, Transition residual, \(P_{\mathrm{mech}}-P_{\mathrm{emp}}\), under the training-defined empirical \(K=6\) partition. \textbf{c}, Validation and confirmation one-step, landscape, local-drift, direction and speed losses; lower values indicate closer agreement. Panels \textbf{a,b} use 3,233,208 confirmation intervals from 56,195 users; panel \textbf{c} compares the validation (\(n=56{,}230\)) and confirmation (\(n=56{,}195\)) cohorts after interval eligibility filtering.}
    \label{fig:minimal_mechanism_kinetics}
\end{figure*}

\subsection*{A parsimonious four-term mechanism recovers the effective dynamics}

We then tested whether a parsimonious conditional one-step mechanism, reinitialised at each observed pre-state, could compress the empirical dynamics within the fixed accounting scaffold.

Within the prespecified hierarchy, the offset dual-channel mechanism was the least complex family meeting the one-standard-error and practical-equivalence criteria (Fig.~\ref{fig:minimal_mechanism_field}a). Its four family-varying terms were a signed response offset, response-restoring feedback, a baseline exposure-alignment drive and a response--support contrast. Its bootstrap score (\(0.2642\)) was practically equivalent to the seven-term reference (\(0.2590\); paired difference \(0.00512\), 95\% interval \(0.00317\)--\(0.00714\)); no refitted one-term deletion remained eligible. Selection was unchanged under equal objective weights and practical-equivalence margins \(0.010\)--\(0.030\), and no family of equal or lower complexity Pareto-dominated it in 300 paired-user bootstraps. Thus, four-term parsimony is supported within the prespecified hierarchy, five-component objective and finite search domain (Supplementary Note~3 and Supplementary Table~5).

After calibration on pooled training and validation data, the four-term specification was fixed and evaluated on 56,195 independent confirmation users and 3,233,208 eligible intervals. Next-state correlations were 0.9380 for response order and 0.9822 for exposure alignment, with interval-weighted root-mean-square errors (RMSEs) of 0.1127 and 0.0359, respectively (Fig.~\ref{fig:minimal_mechanism_kinetics}a); user-balanced next-state occupancy had a JS divergence of 0.1653 (Fig.~\ref{fig:minimal_mechanism_field}b,c). Because level correlations inherit persistence, drift-field recovery was evaluated separately.

Across 954 jointly supported cells, the user-balanced confirmation field closely matched the empirical field: drift-vector \(r=0.9457\) (fixed-support learner-cluster bootstrap 95\% interval, \(0.9351\)--\(0.9546\)), drift-speed \(r=0.9423\) and field RMSE \(0.0471\) (Fig.~\ref{fig:minimal_mechanism_field}c,d; Supplementary Table~11a). This recovered basin-like geometry beyond the marginal next-state distribution. In a no-refit post hoc analysis, the occupancy-weighted distance to the empirical field fell from \(0.0907\) for the construction-null expectation to \(0.0217\) for the mechanism. The corresponding squared-distance reduction was \(0.0078\) (95\% learner-reweighting interval, \(0.0076\)--\(0.0079\)), yielding null-relative skill \(0.943\). The null-subtracted correction also outperformed the validation-fitted coordinatewise rescaling and closely matched the empirical excess in direction and amplitude (weighted local cosine \(0.9473\); amplitude slope \(0.9748\); Supplementary Note~5 and Supplementary Table~9). One-step, landscape and speed losses changed little between validation and confirmation (Fig.~\ref{fig:minimal_mechanism_kinetics}c).

Transitions and residence were excluded from selection, making coarse kinetics an independent post-selection test. Under the training-defined \(K=6\) partition, mean row-wise TV was \(0.1021\), self-transition profiles correlated at \(r=0.9870\), and five of six rows remained diagonal-dominant; \(S_0\) was the sole dominant-edge mismatch (Fig.~\ref{fig:minimal_mechanism_kinetics}b; Supplementary Table~11b). The closure therefore recovered the persistence hierarchy and dominant coarse transition organisation.

Transition-implied geometric residence preserved persistence ordering, but statewise mean-residence ratios of \(0.831\)--\(2.742\) showed how one-step discrepancies amplify over long horizons (Supplementary Table~6). This amplification delineates the closure boundary: conditional one-step flow and persistence ordering are recovered, whereas empirical residence tails retain structure beyond a homogeneous one-step map. Autonomous residence generation was not an estimand because each update was reinitialised from the observed pre-state.

\begin{figure*}
    \centering
    \includegraphics[width=\textwidth]{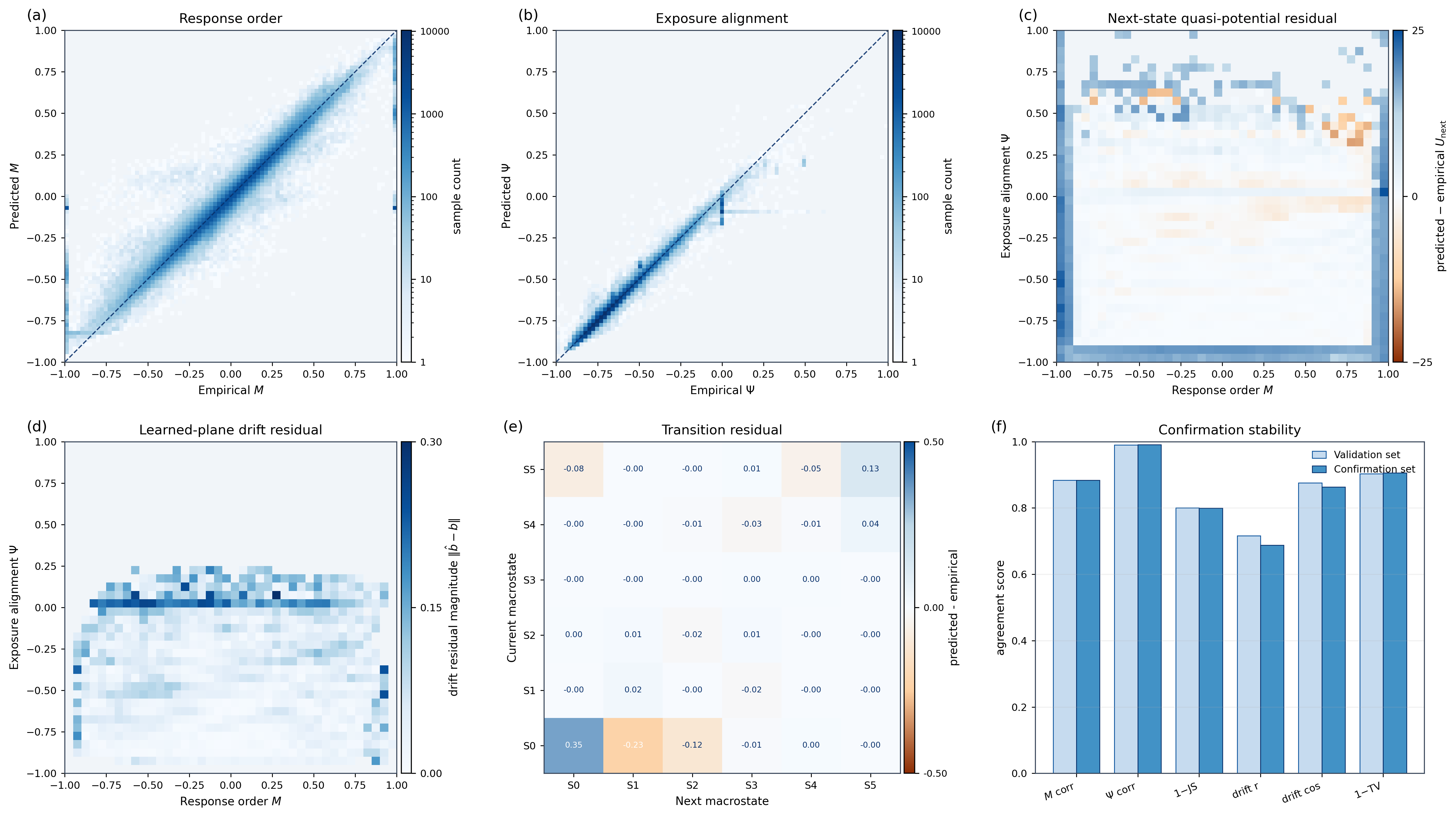}
    \caption{\textbf{Predictive-state Event-SSL recovers effective macrostate dynamics.} \textbf{a,b}, Confirmation densities of empirical and learned current response order \(M\) and exposure alignment \(\Psi\); dashed lines denote identity and colour indicates sample count on a logarithmic scale. \textbf{c}, Predicted-minus-empirical next-state occupancy-derived quasi-potential residual, \(\Delta U_{\mathrm{next}}\). \textbf{d}, Magnitude of the learned-plane drift residual relative to the empirical field, \(\|\hat{\mathbf b}-\mathbf b\|\); pale cells lack common support. \textbf{e}, Learned-minus-empirical transition residual under the training-defined \(K=6\) partition, applied without refitting. \textbf{f}, Validation-to-confirmation stability of coordinate, next-state landscape, learned-plane drift and transition measures; landscape and transition similarities are \(1-\mathrm{JS}\) and \(1-\mathrm{TV}\), respectively. Panels \textbf{a,b} show up to 200,000 confirmation intervals; panels \textbf{c--f} use all eligible intervals in each cohort.}
    \label{fig:event_ssl_macrostate_recovery}
\end{figure*}

\subsection*{Predictive-state Event-SSL recovers effective macrostate dynamics}

We next tested whether predictive-state Event-SSL could recover the phase plane from event histories. Current and next coordinates, increments, evidence maturity and region or mesostate labels were excluded from inputs; \(M,\Psi\) appeared only as auxiliary current-state and one-step targets alongside multi-horizon future prediction. Occupancy, quasi-potential, learned-plane drift and transitions were evaluation-only tests of dynamical self-consistency. Future-prediction-only (pure SSL) and task-only controls tested macrostate accessibility without state or closure targets.

On the held-out confirmation cohort of \(56{,}195\) users and \(3{,}233{,}208\) eligible intervals, the validation-selected model achieved current-state correlations \(r_M=0.8837\) and \(r_\Psi=0.9905\) (RMSEs, \(0.1622\) and \(0.02969\)) and one-step RMSEs of \(0.1100\) and \(0.03535\) (Fig.~\ref{fig:event_ssl_macrostate_recovery}a,b). Under strict user-equal weighting, \(M\) readout fell to \(r_M=0.7046\), whereas \(\Psi\) remained at \(r_\Psi=0.9884\); one-step RMSEs increased to \(0.2990\) and \(0.0935\). This axis-specific sensitivity concerns readout under the user-equal estimand, not the already user-balanced field estimates. Both estimands retained distinct response-order and exposure-alignment information (Supplementary Note~6 and Supplementary Table~10).

Although empirical and predicted next-state occupancies differed at fine scale (JS divergence \(0.2005\)), Event-SSL recovered large-scale quasi-potential geometry that was not directly optimised (Fig.~\ref{fig:event_ssl_macrostate_recovery}c).

Empirical-anchor drift correlated with the empirical field at \(r=0.8802\). With predicted states at both ends, learned-plane correlation was \(r=0.6877\) (fixed-support learner-cluster bootstrap 95\% interval, \(0.6653\)--\(0.7089\)) and occupancy-weighted local cosine was \(0.8629\) (Fig.~\ref{fig:event_ssl_macrostate_recovery}d; Supplementary Table~11a). Coordinate distortion therefore affected global agreement more than local direction. The stricter post hoc empirical-anchor decomposition separated directional recovery from calibration: none of the six models reduced overall field error below either the exact construction-null expectation or the validation-fitted coordinatewise rescaling, and overall and \(M\)-specific skills were negative. Nevertheless, all six retained excess-vector alignment (mean \(r=0.5639\)), partial amplitude (slope \(0.4947\)) and positive \(\Psi\)-specific skill (mean \(0.4660\)). This axis-resolved pattern was consistent across all six seeds, localising the deficit to response-order calibration while preserving a reproducible directional, partial-amplitude component of the excess field (Supplementary Note~5 and Supplementary Table~9, panel~\textbf{d}).

\begin{figure*}
    \centering
    \includegraphics[width=\textwidth]{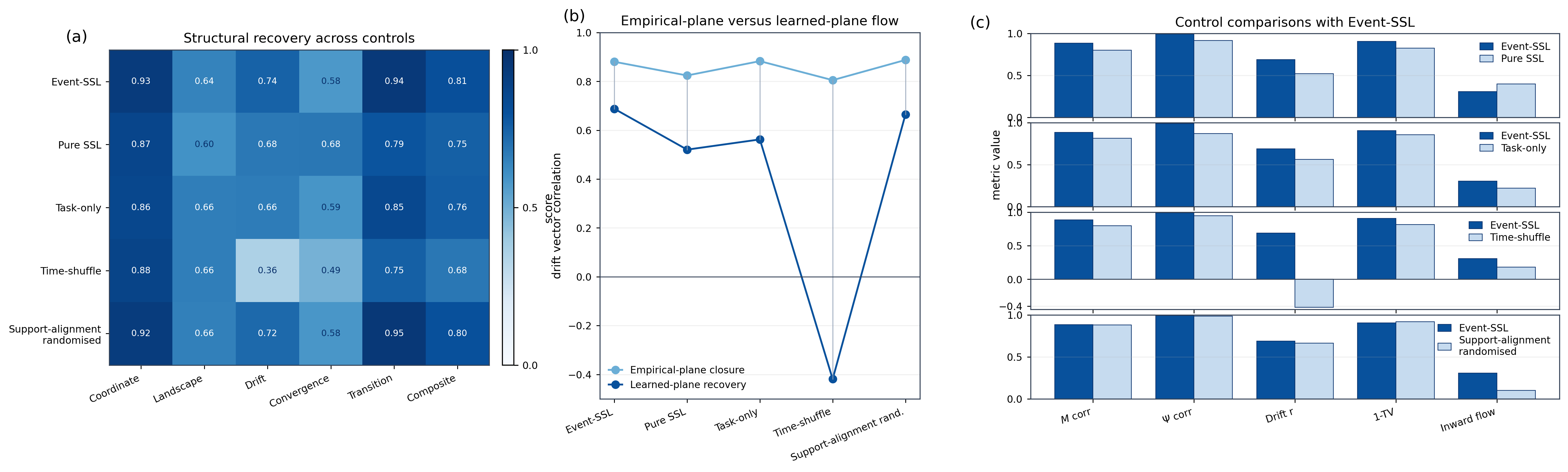}
    \caption{\textbf{Controls dissociate macrostate accessibility from dynamical validity.} \textbf{a}, Confirmation-set descriptive recovery scores for Event-SSL and four controls across coordinates, next-state landscape, learned-plane drift, convergence diagnostics and transitions; the final column is the macrostructure composite. \textbf{b}, Empirical-anchor and learned-plane drift recovery; the former uses the empirical current macrostate and the latter each model's own readout. \textbf{c}, Coordinate recovery, learned-plane drift, transition similarity \(1-\mathrm{TV}\) and inward-flow fraction. Transition metrics use the fixed empirical \(K=6\) partition without refitting. Shuffled-to-ordered transfer is shown on a signed scale to retain its negative learned-plane drift correlation.}
    \label{fig:event_ssl_controls}
\end{figure*}

Using the fixed empirical \(K=6\) partition, the learned and empirical transition matrices had mean row-wise TV \(0.09417\) (Fig.~\ref{fig:event_ssl_macrostate_recovery}e). Dominant outgoing edges matched in all six mesostates, with self-transition largest in every row; the six-mesostate self-transition profiles had descriptive correlation \(r=0.8462\) (Supplementary Table~11b). Because transitions were not training targets, this agreement independently tests coarse kinetic organisation.

For the model selected on validation, all six plotted agreement measures differed by at most \(0.0286\) between validation and confirmation; the same recovery pattern held across all six seeds (Fig.~\ref{fig:event_ssl_macrostate_recovery}f; Supplementary Table~7). Event-SSL therefore recovered macrostate identity, learned-plane flow and coarse mesostate organisation across independent users, supporting representation-level dynamical self-consistency.

\subsection*{Controls dissociate macrostate readout from dynamical validity}

Full Event-SSL was compared with pure SSL, task-only training, within-user time shuffling and support-alignment randomisation. Training-fitted probes supplied readouts for the first two. The temporal control tested shuffled-to-ordered transfer; the alignment control permuted support profiles, including alignment and composition, while preserving responses, temporal order and total support activity. All analyses used the prespecified phase plane and training-defined \(K=6\) partition (Fig.~\ref{fig:event_ssl_controls}a).

The controls distinguished empirical-anchor closure from dynamics generated in the representation's own coordinates (Fig.~\ref{fig:event_ssl_controls}b). Anchor drift correlations remained \(0.805\)--\(0.888\) across models, whereas learned-plane correlations ranged from \(-0.419\) to \(0.688\). The contrast therefore separates empirical-anchor closure accuracy from correctly oriented dynamics in the model's own coordinates.

Substantial macrostate structure persisted without explicit state or closure supervision (Fig.~\ref{fig:event_ssl_controls}a,c). Pure SSL increased inward flow (\(0.396\) versus \(0.306\)) but weakened learned-plane field (\(0.520\) versus \(0.688\)) and self-transition (\(0.331\) versus \(0.846\)) correlations. Task-only lowered next-state JS (\(0.182\) versus \(0.201\)) and increased local cosine (\(0.871\) versus \(0.863\)), yet likewise weakened field (\(0.562\) versus \(0.688\)) and self-transition (\(0.544\) versus \(0.846\)) correlations. Both controls retained two held-out canonical directions. Full Event-SSL therefore showed the strongest joint dynamical consistency without dominating every individual diagnostic (Supplementary Note~4 and Supplementary Table~8, panel~\textbf{e}).

The shuffled-order model preserved coordinate correlations of \(0.795\) and \(0.944\) and empirical-anchor drift correlation \(0.805\), but yielded a globally anticorrelated learned-plane field on ordered trajectories (\(r=-0.419\)). This global anticorrelation replicated in validation (\(r=-0.439\)) and all six seeds (mean, \(-0.434\); 95\% seed-\(t\) interval, \(-0.477\) to \(-0.390\)). Local cosine fell from \(0.863\) to \(0.145\), transition TV rose from \(0.094\) to \(0.187\), and self-transition correlation fell from \(0.846\) to \(0.202\), while inward flow remained positive at \(0.1824\) (Event-SSL, \(0.3059\)). Readout and anchor closure thus survived despite global field anticorrelation, showing that field orientation, unlike state accessibility, was order-sensitive in this transfer test.

Support-alignment randomisation left coordinate recovery nearly unchanged and reduced next-state JS (\(0.189\) versus \(0.201\)) and transition TV (\(0.083\) versus \(0.094\)); learned-plane drift correlation remained similar (\(0.6645\) versus \(0.6877\)). Yet inward flow fell by 66.9\%, from \(0.3059\) to \(0.1012\). This selective dissociation held in validation and all six seeds: basin-directed transport was more sensitive to support alignment than state location or coarse-transition recovery.

Across seeds, shuffled-order learned-plane field correlation was always negative and support-alignment randomisation always reduced inward flow. Full Event-SSL exceeded task-only in all six and pure SSL in five of six field-correlation comparisons (Supplementary Table~7).

\begin{figure*}
    \centering
    \includegraphics[width=\textwidth]{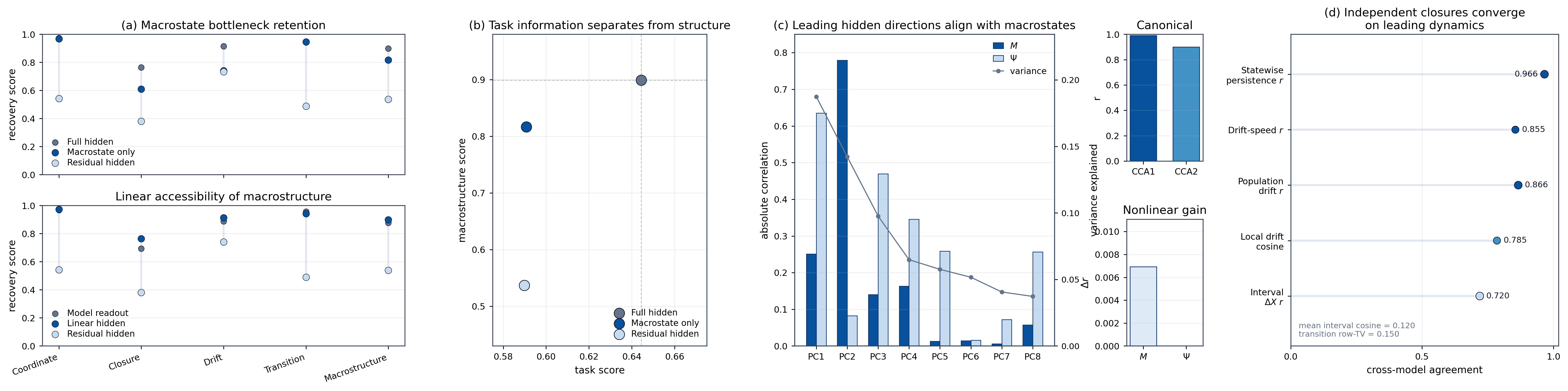}
    \caption{\textbf{A low-dimensional effective state organises the hidden representation.} \textbf{a}, Confirmation recovery from the full hidden state, two-coordinate bottleneck and residual hidden activity (top), and from the trained readout, a training-fitted linear readout and its residual (bottom). \textbf{b}, Response-task versus macrostructure recovery. \textbf{c}, Leading-PC alignment with empirical coordinates, training variance explained, canonical correlations and nonlinear-over-linear probe gain. \textbf{d}, Multi-scale agreement between the independently fitted mechanism and Event-SSL on matched confirmation intervals; points report self-transition, drift-speed, field, displacement and local-cosine agreement, and annotations report interval cosine and transition TV. Field metrics use 954 jointly supported cells. Both closures use matched empirical anchors and the training-defined \(K=6\) partition without cross-model fitting.}
    \label{fig:event_ssl_representation}
\end{figure*}

\subsection*{A low-dimensional effective state organises the hidden representation}

Because controls without state or closure objectives retained macrostate information, we tested whether the state organised the recurrent representation beyond the output head. After training, readouts and residualisation maps fitted on training data were applied unchanged to validation and confirmation. We compared the full hidden state, a two-coordinate bottleneck with a training-fitted quadratic closure, and residual activity after subtracting the bottleneck-predictable component (Fig.~\ref{fig:event_ssl_representation}).

On confirmation, the bottleneck retained nearly all coordinate and transition-domain scores and about four-fifths of closure and drift scores, yielding 90.8\% of the full-hidden descriptive macrostructure composite (Fig.~\ref{fig:event_ssl_representation}a). Raw field recovery was scale-dependent: learned-plane \(r=0.565\) and local cosine \(0.401\), versus \(0.717\) and \(0.942\) for the full hidden state. Within-user permutations placed local direction and coarse transition topology above their model-specific floors, but not global field correlation or statewise persistence. The two-dimensional component is therefore robust locally and topologically, not a complete dynamical description (Supplementary Table~8, panel~\textbf{c}).

Post hoc state-only closures improved matched-origin drift in all six seeds and reached transition TV \(0.051\) with self-transition \(r=0.993\), but transferred incompletely to the primary learned-plane gauge. Both current-state readout mismatch and conditional-closure error therefore contribute (Supplementary Table~8, panel~\textbf{d}).

Residualisation revealed complementary information. The residual representation retained 59.7\% of the full-hidden descriptive macrostructure score but 80.2\% of the descriptive drift score, indicating that higher-dimensional directions preferentially refined local motion. Yet the bottleneck and residual retained nearly identical response-task scores (91.7\% and 91.5\% of full hidden), showing that task information was distributed across both components (Fig.~\ref{fig:event_ssl_representation}b).

A training-fitted linear projection exceeded the trained readout in confirmation macrostructure score (\(0.8985\) versus \(0.8767\)); residualisation reduced it to \(0.5391\) (Fig.~\ref{fig:event_ssl_representation}a). The macrostate was therefore broadly linearly accessible rather than confined to the decoder.

Across six seeds, the two canonical correlations averaged \(0.991\) and \(0.900\) for full Event-SSL, \(0.930\) and \(0.817\) for pure SSL, and \(0.870\) and \(0.810\) for task-only. Both controls therefore retained two-axis access without direct macrostate or closure supervision. In the full model, the first two principal components explained 33.0\% of hidden variance and aligned preferentially with \(\Psi\) and \(M\) (Fig.~\ref{fig:event_ssl_representation}c); control access was less concentrated in leading variance directions and showed larger feature-matched nonlinear gains. The full objective was therefore associated with a more prominent, nearly linear macrostate that remained accessible in both controls (Supplementary Table~8, panel~\textbf{e}).

Across cohorts and seeds, the two-coordinate state remained a leading but non-exhaustive component: it captured state identity, one-step closure and coarse transition topology, while residual directions retained complementary drift and task information (Supplementary Table~7).

\subsection*{Neural and mechanistic closures share leading effective dynamics}

We compared the independently fitted neural and four-term closures across interval, population and mesostate scales on all \(3{,}233{,}208\) matched confirmation intervals from \(56{,}195\) users (Fig.~\ref{fig:event_ssl_representation}d). Both were evaluated against the same held-out empirical dynamics, using numerically matched current-state anchors and next-state targets under the training-defined \(K=6\) partition, with no updating on confirmation data or cross-model fitting.

Agreement increased with coarse-graining. Next-state predictions were strongly correlated (\(r=0.9238\) for \(M\), \(0.9779\) for \(\Psi\)), whereas interval displacements were only partly aligned (\(r=0.7196\); mean cosine, \(0.1201\)). State-space averaging raised agreement to drift-vector \(r=0.8661\) across 954 cells (fixed-support learner-cluster bootstrap 95\% interval, \(0.8569\)--\(0.8747\)), drift-speed \(r=0.8552\) and weighted local cosine \(0.7851\) (Fig.~\ref{fig:event_ssl_representation}d; Supplementary Table~11a). Conditioning on the shared empirical target attenuated but did not eliminate the six-seed association (partial \(r=0.2048\); 95\% seed-\(t\) interval, \(0.1920\)--\(0.2177\)); empirical-error vectors remained positively aligned globally (mean \(r=0.2140\)) but not locally in direction. The closures therefore share a leading population field under a common empirical gauge, not pointwise corrections.

The two models were strongest at different scales. When evaluated from the same empirical current-state anchors, Event-SSL tracked empirical interval displacements better than the mechanism (\(r=0.7551\) versus \(0.6062\)). The mechanism recovered the population field more closely (\(r=0.9457\) versus \(0.8802\)) and gave lower next-state occupancy divergence (\(0.1653\) versus \(0.2005\)) and empirical-anchor transition row-wise TV (\(0.1021\) versus \(0.1512\)). Event-SSL captured more interval-specific variation, whereas the four-term closure recovered the more accurate low-order population field.

Cross-model convergence was selective: it was strongest in restoring flow and persistence hierarchy, not density mass or off-diagonal routes. Model occupancies differed by Jensen--Shannon divergence \(0.2096\), and transition matrices by mean row-wise TV \(0.1497\) (learner-cluster bootstrap 95\% interval, \(0.1447\)--\(0.1547\)), yet the six-mesostate self-transition profiles had descriptive Pearson correlation \(r=0.9660\), identical rank ordering (\(\rho=1.000\)) and leave-one-state-out correlations of \(0.9608\)--\(0.9841\). Both also followed the empirical persistence hierarchy. The four-term mechanism therefore provides a compact surrogate for Event-SSL's leading restoring flow and persistence hierarchy, while density mass and off-diagonal routes remain model-specific (Supplementary Table~11b,c).

\section*{Discussion}

At the submitted-bundle scale, the semantic state \(\mathbf X=(M,\Psi)\) supports reproducible basin-like flow and, under the stated Kaplan--Meier censoring assumption, operationally defined, state-heterogeneous metastable-like kinetics. The construction-matched null showed that normalised-memory accounting contributes generic relaxation but does not determine the replicated excess field: full-field and negative-divergence departures reproduced, whereas uniform core slowing did not. On independent confirmation users, the family-bounded four-term mechanism recovered the field with \(r=0.946\) (fixed-support learner-cluster bootstrap 95\% CI, \(0.935\)--\(0.955\)) and null-relative skill \(0.943\); Event-SSL recovered the same state, self-consistent learned-plane dynamics and directional, partial-amplitude and \(\Psi\)-specific excess structure without full calibration. Thus, closed-loop behavioural logs can support empirically testable and compressible macroscopic dynamics rather than only predictive regularities \cite{Barnett2023,Lucke2024,Mardt2018,Wang2019,Gao2024Dynamics,Moore2025}.

The controls show that macroscopic recovery is hierarchical rather than binary. Pure SSL and task-only retained two macrostate directions but did not match Event-SSL's joint field and transition coherence. Shuffled-order training preserved readout while reversing learned-plane flow in all six seeds; support-alignment randomisation preserved state location and coarse transitions but reduced inward transport by \(66.9\%\) for seed 42, with a reduction in every seed. Static accessibility, field orientation and basin-directed transport are thus separable representation properties. These controls identify representation-level dependencies, not interventions on learners.

Event-SSL captured interval displacement more closely, whereas the four-term mechanism recovered the population field more accurately. Their agreement strengthened under coarse-graining (field \(r=0.866\); identical persistence rank) without cross-model fitting, while density mass, off-diagonal routes and local residual direction remained model-specific. Empirical residence tails also retained structure beyond a homogeneous one-step map, separating conditional state-update closure from longer-horizon duration structure. Within Event-SSL, the two-coordinate bottleneck retained \(90.8\%\) of descriptive macrostructure, while residual activity retained \(80.2\%\) of drift structure and comparable task information. The state is therefore a leading organising backbone rather than an exhaustive dynamical description \cite{Machta2013,Transtrum2014,Transtrum2015,Fabiani2024,Recanatesi2021}.

Because the coordinates were defined from event meaning before fitting, their empirical status rests on held-out field replication, construction-aware excess, low-order closure and approximately linear accessibility. Field geometry remained stable across memory and activity variants (\(r=0.962\)--\(0.981\)), and the alignment-based state showed approximately threefold greater null-referenced prominence than a denominator-matched activity--idle comparator in validation and confirmation. This single-comparator result does not establish coordinate optimality or causal necessity, but it shows that reproducible contraction alone does not explain the stronger excess exposed by the alignment-based gauge. The central advance is an externally anchored effective field recovered independently from empirical trajectories, a compact mechanism and a predictive representation \cite{Gao2024Dynamics,Moore2025,Recanatesi2021,Machta2013,Kornblith2019,Huh2024,Beretta2025,Simon2026}.

The transferable contribution is the identification procedure rather than the numerical form of \(M\) and \(\Psi\): prespecify semantic observables, test held-out dynamics against construction-aware expectations, compare matched gauges, seek low-order closure and test coherent evolution in predictive representations. The evidence is scoped to one adaptive platform and conditional event-time closure; population-level reproducibility does not imply learner-level homogeneity. Within that scope, the procedure provides a falsifiable template for closed-loop systems in which agent actions reshape future observations, including adaptive platforms, co-adaptive interfaces and embodied agents \cite{Bartolozzi2022,Hafner2025,Madduri2026}.

\section*{Methods}

All analyses used a common submitted-bundle event-time panel and prespecified \((M,\Psi)\) phase plane. Training defined response baselines, convergence criteria and mesostate centres; validation assessed replication and guided development; confirmation remained held out until all specifications were final. Auxiliary variables entered interval construction, model inputs or diagnostics without redefining the primary state.

\subsection*{Event panel and macrostate construction}

We processed the public EdNet-KT4 logs and associated question, bundle, explanation, lecture, payment and coupon metadata \cite{Choi2020}. Action names were standardised; records with invalid timestamps, unrecognised actions, malformed item identifiers or exact duplicates were removed; and events were stably ordered by timestamp, within-timestamp action priority and item identifier. Final responses, bundle attempts and study episodes were reconstructed per user. Choice revisions and media interactions served as process measurements; payment, refund and coupon events provided access context only.

Submitted bundles defined event time. The pre-state occurred at bundle entry or, when entry was unavailable, at submission. The interval update comprised final responses in that bundle and support activity from submission to the next submitted-bundle pre-state. A random user partition generated with seed 42 assigned all 297,915 users to training (\(n=178{,}749\)), validation (\(n=59{,}583\)) and confirmation (\(n=59{,}583\)) cohorts. Analyses requiring a defined state or an adjacent observation consequently included fewer observations.

For question \(q\) with mapped tags \(\mathcal T_q\), each final response assigned mass \(r_{qi}=|\mathcal T_q|^{-1}\) to \(i\in\mathcal T_q\). Let \(n_i,c_i\) denote tag-level response and correct-response mass, \(n_q,c_q\) question counts and \(p_0\) the global mapped-response baseline estimated from the training cohort. The corresponding shrinkage baselines were

\[
\begin{aligned}
p_i^{\mathrm{tag}}
  = \frac{c_i+20p_0}{n_i+20},\qquad
p_q
  = \frac{c_q+50p_q^{\mathrm{parent}}}{n_q+50}.
\end{aligned}
\]

Here \(p_q^{\mathrm{parent}}\) was the mean tag baseline for \(q\), or \(p_0\) when no mapped-tag baseline was available. The resulting baselines were clipped to \([10^{-3},1-10^{-3}]\) and then held fixed; questions without mapped tags did not update the response-evidence accumulators.

For user \(u\), final response \(Y_{uqm}\in\{0,1\}\) in bundle \(m\) contributed

\[
\begin{aligned}
s_{uqmi} = r_{qi}(Y_{uqm}-p_q),\qquad
a_{uqmi} = r_{qi}|Y_{uqm}-p_q|.
\end{aligned}
\]

Tagwise accumulators \(S_{ui,n}\) and \(A_{ui,n}\) contained preceding contributions and decayed between pre-states as \(\exp[-\max(\Delta t,0)/(10\,\mathrm{days})]\). Response order and unresolved demand were

\[
\begin{aligned}
M_{un}
  = \frac{\sum_i S_{ui,n}}{\sum_i A_{ui,n}},\qquad
D_{ui,n}
  = \max\{-S_{ui,n},0\}.
\end{aligned}
\]

Only final-answer correctness updated these accumulators, and \(M_{un}\) was defined when the evidence mass was positive.

Exposure alignment used a separate 10-day memory of aligned \(H^+\), off-target \(H^-\), neutral-active \(H^0\) and idle \(I\) mass. For mapped activity, cosine similarity to the relevant demand vector was clipped to \([0,1]\) and used as the aligned fraction; undefined alignment was neutral. Current-bundle response activity was compared with pre-state unresolved demand, whereas mapped support observed after submission was compared with pre-state demand plus current-bundle content. With \(g_h(t)=t/(t+h)\), \(f_{\mathrm{answered}}\) denoting the fraction of bundle questions answered and \(t_{\mathrm{response}}\) the bundle response duration, response activity was

\[
\alpha_n^{\mathrm{resp}}
  =
  \max\!\left\{
  f_{\mathrm{answered}},
  \tfrac12 g_{3\,\mathrm{min}}(t_{\mathrm{response}})
  \right\}.
\]

Explanation and lecture activity used the larger dwell- or media-based proxy, capped at one; duration-based proxies used half-saturation times of \(2.5\) and \(4\) min, respectively. Unmapped support was neutral; uncovered duration contributed \(g_{1\,\mathrm{day}}(t_{\mathrm{idle}})\) to idle mass. Response activity and final-response evidence preceded support and idle updates. Exposure alignment was

\[
\Psi_{un}
  =
  \frac{H^+_{un}-H^-_{un}}
       {H^+_{un}+H^-_{un}+H^0_{un}+I_{un}}.
\]

The pre-state was \(\mathbf X_{un}=(M_{un},\Psi_{un})^\top\), and \(\Delta\mathbf X_{un}=\mathbf X_{u,n+1}-\mathbf X_{un}\). Final pre-states contributed to occupancy but not drift; missing steps were not interpolated.

\subsection*{Empirical fields and kinetic diagnostics}

The \((M,\Psi)\) plane used a \(40\times40\) grid on \([-1,1]^2\), with the upper boundary included in the final bin. Let \(N_u\) be the number of submitted-bundle observations for user \(u\) in the relevant cohort; each observation received \(w_{un}=N_u^{-1}\) before eligibility filtering. Thus, each learner had unit total weight on the full panel, while its retained occupancy and drift masses were \(N_u^{S}/N_u\) and \(N_u^{T}/N_u\), where \(N_u^{S}\) and \(N_u^{T}\) denote its finite in-range state and transition counts. We refer to this as full-panel learner-normalised weighting; the strict sensitivity instead used \(1/N_u^{S}\) and \(1/N_u^{T}\) over eligible rows. Only finite states entered occupancy and finite increments entered drift. Cell occupancy \(\widehat p_c\) was the normalised weight sum; the quasi-potential \(U_c=-\log(\widehat p_c+10^{-12})\) was used as a descriptive occupancy landscape.

Conditional event-time drift and increment covariance were

\[
\begin{aligned}
\widehat{\mathbf b}_c
  &= \mathbb E_w[
     \Delta\mathbf X_{un}\mid\mathbf X_{un}\in c],\\
\widehat{\boldsymbol\Sigma}_c
  &= \operatorname{Cov}_w[
     \Delta\mathbf X_{un}\mid\mathbf X_{un}\in c].
\end{aligned}
\]

Drift cells required 30 transitions. Differential quantities required a complete five-cell stencil, with at least 50 states, five users and supported drift in every cell.

The convergence core was defined from the training field. On supported stencils, candidate cells had divergence at or below the 0.80 quantile of negative values, while drift speed and

\[
R_c
  =
  \frac{\|\widehat{\mathbf b}_c\|_2}
       {\sqrt{\operatorname{tr}(\widehat{\boldsymbol\Sigma}_c)}+10^{-12}}
\]
were at or below their respective 0.60 quantiles. Eight-neighbour components of at least four cells qualified when at least half of the flow-weighted support in a radius-\(0.35\) shell pointed inward, the core-to-shell speed ratio was below one and the flow-weighted mean divergence was negative. Among qualifying components, the primary core was selected by a flow-only convergence score, followed by user-balanced drift support and supported-cell count. The selected region, shell radius and thresholds were then held fixed; occupancy entered neither selection, ranking nor centring.

To test construction-implied relaxation, we used a construction-matched accounting null. Let \(M=S/E\) and \(\Psi=G/B\), where \(S=ME\), \(B=H^++H^-+H^0+I\) and \(G=\Psi B\). For each transition,
\[
\begin{aligned}
(A_n^M,J_n^M)
&= (E_n^{\mathrm{resp}}-E_n, S_n^{\mathrm{resp}}-S_n), \\
\qquad
(A_n^\Psi,J_n^\Psi)
&= (B_n^{\mathrm{post}}-B_n, G_n^{\mathrm{post}}-G_n),
\end{aligned}
\]
with \(Z_n^x=J_n^x/A_n^x\) for \(A_n^x>10^{-12}\), otherwise zero. Here, the superscripts \(\mathrm{resp}\) and \(\mathrm{post}\) denote the response-updated and post-interval accumulator states, respectively, before the common decay. Across 100 replicates, the joint pairs \((Z_n^M,Z_n^\Psi)\) were permuted without fixed points within disjoint opportunity-matched strata; cutpoints for continuous matching variables were estimated from the training cohort. Each replicate preserved the observed current states, denominator increments, user-balanced weights, empirical grid support and the training-defined core and shell. Null next states were
\[
M_{n+1}^{*}
=
\frac{S_n+A_n^M\widetilde Z_n^M}
     {E_n+A_n^M},
\qquad
\Psi_{n+1}^{*}
=
\frac{G_n+A_n^\Psi\widetilde Z_n^\Psi}
     {B_n+A_n^\Psi},
\]
with common decay cancelling from both ratios. The resulting coordinates were bounded to \([-1,1]\). The primary test used the occupancy-weighted root-mean-square drift-field distance with leave-one-out null means and the \(+1\) Monte Carlo correction. Negative-divergence occupancy, shell inward flow and the core-to-shell speed ratio used prespecified one-sided tests with Benjamini--Hochberg adjustment \cite{Benjamini1995}. Prespecified inference was conducted in the validation cohort, and the confirmation cohort provided an independent replication using the same coordinate definition, grid, support thresholds, convergence core and null construction (Supplementary Note~1).

Continuous-field analyses did not use clustering. For kinetic summaries, \(K=6\) was fixed a priori. \(k\)-means was fitted to standardised training coordinates with 20 initialisations, using a sample of at most 500,000 states drawn with random seed 42 under user-balanced sampling and weighting \cite{MacQueen1967,Pedregosa2011}. The standardisation parameters, centres ordered by \(M\) and then \(\Psi\), and state labels derived from the training cohort were applied unchanged to held-out observations and model predictions. Transitions used unweighted, row-normalised counts of within-user steps with consecutive bundle indices; the partition served only as an operational kinetic coarse-graining.

Residence episodes used the same partition. An exit was observed only when the next step entered another mesostate; user boundaries, gaps in the submitted-bundle sequence and unobserved mesostates right-censored the episode. Kaplan--Meier inference therefore assumes non-informative right censoring within each mesostate. Statewise Kaplan--Meier estimates \cite{KaplanMeier1958} were compared with \(S_i^{\mathrm{geo}}(\ell)=P_{ii}^{\,\ell-1}\). Restricted mean residence time (RMST) was evaluated up to the largest duration with at least 20 episodes at risk. Prespecified upper-tail inference was restricted to \(\ell=10\), using a one-sided normal approximation based on Greenwood variance for the Kaplan--Meier estimator, conditional on empirical \(P_{ii}\), with Benjamini--Hochberg adjustment across six mesostates \cite{Greenwood1926,Benjamini1995}; full-curve departures were descriptive. These are population-level statewise estimands under the observed learner mixture. Excess over the homogeneous geometric reference may therefore reflect within-learner duration dependence, persistent between-learner heterogeneity or both; either source places the aggregate kinetics beyond a homogeneous statewise Markov description.

Kinetic robustness analyses used only the validation cohort and were conducted independently of the mechanistic and neural models. A recursive surrogate propagated the state within each contiguous observable segment while preserving the opportunity-matching strata, denominator paths and censoring structure. Using the prespecified 10-step horizon, a post hoc equal-state mean log RMST-lift endpoint was tested against 100 surrogates with the \(+1\) correction. A complementary post hoc studentised maxT test evaluated \(D_{10,i}=\widehat S_i(10)-P_{ii}^{9}\) across six states; positivity required \(D_{10,i}>0\) and family-wise-error-adjusted \(p<0.05\). These endpoints test all-state integrated and state-specific fixed-horizon excess, respectively. Fixed \(K=4,5,7,8\) partitions tested resolution, and a 1,000-replicate learner-cluster bootstrap recomputed transition and Kaplan--Meier quantities at fixed horizons \cite{Efron1979} (Supplementary Note~2 and Supplementary Table~4).

Training-to-validation replication was quantified by occupancy Jensen--Shannon (JS) divergence \cite{Lin1991} and agreement in drift, transitions and residence. Observation-gap sensitivity compared the full panel with analyses restricted to observations having an observed successor or to non-terminal intervals with gaps no longer than 7 days, without redefining the state, convergence core or \(K=6\) partition (Supplementary Table~2).

\subsection*{Minimal effective-mechanism inference}

The empirical state and submitted-bundle interval structure were fixed before mechanism inference. Each observed pre-state initialised an independent, non-recursive one-step update. Let \(E_n\) be response-evidence mass, \(S_n=M_nE_n\) its signed numerator, \(B_n\) the exposure-memory denominator and \(G_n=\Psi_nB_n\) its signed numerator. Each interval supplied an answered-count proxy \(q_n\) derived from bundle and response records; aligned-plus-off-target response and support masses \(A_n^R,A_n^S\); neutral or unmapped masses \(A_n^{R0},A_n^{S0}\); idle mass \(I_n\); and elapsed time \(\Delta t_n\). Memory decay was

\[
\rho_x(n)
=
\exp\!\left[-\frac{\max(\Delta t_n,0)}{\tau_x}\right],
\qquad x\in\{R,A\},
\]
with \(\tau_R=\tau_A=10\) days. Define
\(\Pi(z)=\min\{1,\max(-1,z)\}\) and
\(C_b(z)=\min\{b,\max(-b,z)\}\). The response channel was

\[
\begin{aligned}
R_n
  &= \lambda_R r q_n,\\
U_n
  &= \gamma_R R_n
     \tanh(\theta_0-\theta_M M_n),\\
E_{n+1}
  &= \rho_R(n)(E_n+R_n),\\
S_{n+1}
  &= \rho_R(n)(S_n+U_n),\\
\widehat M_{n+1}
  &= \Pi(S_{n+1}/E_{n+1}).
\end{aligned}
\]

Here \(r\) is residual evidence per answered question, and \(\gamma_R\) is the signed-response calibration gain. Evidence maturity \(V_n=1-\exp(-E_n/\eta)\) was used only for evidence-mass reconstruction, calibration and diagnostics, with \(\eta\) estimated from valid \((E_n,V_n)\) pairs. With \(A_n^{\mathrm{act}}=A_n^R+A_n^S+A_n^{R0}+A_n^{S0}\), the signed exposure increment was

\[
H_n
=
\gamma_A\lambda_A
\left[
A_n^R\tanh(\phi_0)
+
A_n^S\tanh(\phi_0+\delta_S)
\right],
\]

where \(\gamma_A\) is the exposure-alignment calibration gain. The exposure update was

\[
\begin{aligned}
\widetilde B_n
  &= B_n+\lambda_A A_n^{\mathrm{act}},\\
\widetilde G_n
  &= C_{\widetilde B_n}(G_n+H_n),\\
B_{n+1}
  &= \rho_A(n)(\widetilde B_n+\lambda_I I_n),\\
G_{n+1}
  &= \rho_A(n)\widetilde G_n,\\
\widehat\Psi_{n+1}
  &= \Pi(G_{n+1}/B_{n+1}).
\end{aligned}
\]

Unsigned masses were non-negative; ratios were clipped to \([-1,1]\), and \(\widehat M_{n+1}=0\) at vanishing evidence mass. The four family-varying substantive parameters were response offset \(\theta_0\), restoring coefficient \(\theta_M\), alignment baseline \(\phi_0\) and response--support contrast \(\delta_S\). The seven-term reference family included \(\theta_\Psi\Psi_n+\theta_{M\Psi}M_n\Psi_n\) in the response drive and \(-\phi_\Psi\Psi_n\) in both alignment drives; these terms were zero in the selected family. Scales \(\lambda_R=0.46\), \(\lambda_A=1.10\) and \(\lambda_I=0.85\) were fixed across families and excluded from the substantive parameter count.

For family comparison, \(r\) was the median positive decay-adjusted evidence increment per answered question, clipped to \([0.02,2]\); \(\gamma_R\) and \(\gamma_A\) were the channelwise 75th percentiles of the signed-fraction magnitudes, clipped to \([0.1,1]\). These training-derived quantities were fixed during comparison and excluded from the substantive parameter count.

The prespecified hierarchy comprised state persistence, single-channel baselines, a two-coordinate core, response-offset and dual-channel families, coupling extensions and a seven-term reference. Single-channel baselines retained the common accounting scaffold but removed one signed drive. Candidate parameters were screened on prespecified grids in a training subset, refined, and then evaluated in the training and validation cohorts; terms absent from a family were fixed at zero.

Family comparison used

\[
\begin{aligned}
\mathcal L_{\mathrm{primary}}
={}&
0.10\,\mathcal L_{\mathrm{step}}
+0.20\,\mathcal L_{\mathrm{JS}}
+0.30\,\mathcal L_{\mathrm{local}}\\
&+
0.20\,\mathcal L_{\mathrm{dir}}
+0.20\,\mathcal L_{\mathrm{mag}}.
\end{aligned}
\]

The components measured interval-weighted variance-normalised one-step error, user-balanced next-state occupancy JS divergence, local error between user-balanced drift fields, field-correlation loss \((1-r_{\mathrm{field}})/2\) and drift-speed discrepancy. Parameter fitting also penalised phase and activity-coverage discrepancies; family ranking and parsimony used only the five-component score. Convergence, transition, residence and maturity diagnostics were excluded from selection.

Uncertainty used 300 paired validation-user bootstrap replicates \cite{Efron1979}, with the same resampled users applied to every family. The best family minimised mean primary score. Eligibility required a mean within one bootstrap standard error of the best and a 97.5th-percentile paired difference no greater than the prespecified margin 0.02. The simplest eligible family was selected, with the score used to break ties, as an adapted one-standard-error rule \cite{Hastie2009}; margins 0.010--0.030 were also examined.

Direct one-parameter deletions were refitted until no simpler family qualified. Final selection required a non-equivalent persistence baseline and no unresolved boundary optimum. Post hoc sensitivity analyses using only training and validation data repeated the complete selection under equal component weights and assessed Pareto dominance across the five losses and parameter count; confirmation data were not used. Minimality is therefore bounded by the hierarchy, objective, fixed scales and finite search domain.

The selected family was tuned on pooled training and validation data by deterministic hierarchical grid search. Within the near-optimal plateau satisfying the saturation criterion, we selected the smallest finite response--support contrast, yielding

\[
(\theta_0,\theta_M,\phi_0,\delta_S)
=
(-0.17,0.64,-1.35,6).
\]

Calibration on pooled data gave \((\eta,r,\gamma_R,\gamma_A)=(20,0.3865,1,0.9537)\); the family, coefficients and calibration were then fixed before confirmation.

Confirmation used the selected specification without further search, recalibration or spatial redefinition. Diagnostics comprised one-step error, next-state occupancy divergence, local drift error, field direction and drift speed. Post-selection kinetics used the \(K=6\) partition. Residence references were geometric curves implied by the one-step transition matrix, not autonomous mechanism trajectories.

\subsection*{Predictive-state representation learning}

Event-SSL used causal submitted-bundle interval sequences with finite current and next \(M,\Psi\) targets. Inputs comprised response, support, timing, process and access context plus content and delivery identifiers. Current and next coordinates, their increments, evidence maturity and graph-, region- or mesostate-derived variables were excluded. \(M,\Psi\) appeared only as current-state and one-step targets alongside multi-horizon future prediction; spatial and kinetic statistics were used only for evaluation.

Numeric variables were mean-imputed and standardised with training statistics. Selected non-negative count, mass, duration and gap fields were represented both on their original scale and as \(\log(1+x)\); missing transformed values were set to zero before standardisation. Each categorical field was hashed into 32,768 buckets and embedded in 16 dimensions \cite{Weinberger2009}. Sequences were split at user changes or non-contiguous bundle indices.

The numeric block and categorical embeddings were passed through a two-layer 224-dimensional projection with layer normalisation, Gaussian error linear unit (GELU) activations and dropout, followed by a two-layer gated recurrent unit (GRU) with 320 hidden units \cite{Cho2014,Ba2016,Hendrycks2016,Srivastava2014}. For interval embedding \(\mathbf z_n\) and recurrent states immediately before and after the interval, \(\mathbf h_n^{-}\) and \(\mathbf h_n^{+}\),

\[
\begin{aligned}
\mathbf h_n^{+}
  &= \operatorname{GRU}(\mathbf z_n,\mathbf h_n^{-}),\\
\widehat{\mathbf X}_n
  &= \tanh g_{\mathrm{s}}(\mathbf h_n^{-}),\\
\mathbf d_n
  &= 0.5\,\tanh
     g_{\Delta}([\mathbf h_n^{-},\mathbf z_n]),\\
\widehat{\mathbf X}_{n+1}
  &= \tanh(\widehat{\mathbf X}_n+\mathbf d_n),\\
\widehat{\mathbf x}^{\mathrm{num}}_{n+k}
  &= W_k\mathbf h_n^{+}+\mathbf b_k,
  \qquad k\in\{1,2,4\}.
\end{aligned}
\]

Here \(\widehat{\mathbf X}=(\widehat M,\widehat\Psi)^\top\), and \(g_{\mathrm{s}}\) and \(g_{\Delta}\) were two-layer GELU readouts with dropout. The current-state readout used only preceding history; the current interval entered the bounded next-state proposal and recurrent update. Linear future predictors reconstructed the standardised numeric vector.

Training minimised

\[
\mathcal L
=
\mathcal L_{\mathrm{future}}
+0.5\,\mathcal L_{\mathrm{state}}
+0.5\,\mathcal L_{\mathrm{closure}}.
\]

The future term averaged feature-wise squared error across horizons 1, 2 and 4 over pairs observed at both source and target positions; padding and horizons without valid pairs were excluded. State and closure used the smooth-\(L_1\) loss (\(\beta=0.05\)), a scaled Huber form \cite{Huber1964}.

Training used 256-position windows with stride 128. Future loss used all valid pairs; state and closure supervision used the first window of each contiguous sequence after an eight-position warm-up. Training ran for eight epochs with batch size 192, dropout 0.10 and seed 42. AdamW used a learning rate of \(2\times10^{-4}\), weight decay \(10^{-4}\) and gradient clipping at one \cite{Loshchilov2019}; implementation used PyTorch \cite{Paszke2019}. Parameters from the epoch with the lowest validation state-plus-closure loss were retained; confirmation data were excluded from normalisation and model selection. Five additional seeds (\(2026\), \(666\), \(606\), \(37\) and \(4669\)) repeated the full training and evaluation procedure. Student-\(t\) intervals summarise variation across runs rather than learner-population uncertainty.

\subsubsection*{Dynamical recovery}

Validation and confirmation retained sequence order, with recurrent state carried through each contiguous sequence. Coordinate recovery compared model readouts directly with \(M\) and \(\Psi\). Occupancy and drift used the same \(40\times40\) user-balanced \((M,\Psi)\) grid and 30-transition support threshold; quasi-potential was derived from occupancy. We distinguished

\[
\begin{aligned}
\widehat{\Delta\mathbf X}^{\mathrm{anchor}}_n
  &= \widehat{\mathbf X}_{n+1}-\mathbf X_n,\\
\widehat{\Delta\mathbf X}^{\mathrm{learned}}_n
  &= \widehat{\mathbf X}_{n+1}
     -\widehat{\mathbf X}_n.
\end{aligned}
\]

The anchor view used the observed current state; the learned-plane view used model readouts at both ends. Commonly supported cells supplied drift-vector correlation, local drift RMSE, drift-speed correlation, local cosine and occupancy-weighted local cosine. Inward flow used a separate training-defined high-occupancy supported quasi-potential minimum rather than the empirical flow-defined core. Divergence was descriptive and used central differences only on fully supported five-cell stencils.

Transition recovery used the fixed empirical \(K=6\) partition. Training-derived standardisation, ordered centres and state labels were applied unchanged; neither \(k\)-means nor \(K\) was refitted or reselected. Anchor transitions paired observed current with predicted next states, whereas learned-plane transitions used predictions at both ends. Matrices used unweighted, row-normalised counts. Composite scores were descriptive; component metrics supported the scientific comparisons.

\subsubsection*{Structural controls}

The future-prediction-only (pure SSL) control retained the encoder but set the state and closure loss weights to zero; model selection used validation future loss. Ridge-regression probes (\(\alpha=1\)), fitted on at most 300,000 training states, mapped pre- and post-interval hidden activity to the current and next macrostates \cite{Hoerl1970}.

The task-only control used the same causal encoder and binary cross-entropy against clipped current-bundle accuracy. Its head read \(\mathbf h_n^{-}\) before the current interval entered the recurrent update. Initial and continuation windows were supervised after the eight-step warm-up; validation task loss selected the model, and equivalent probes supplied structural readouts.

For the temporal control, intervals were permuted within each user and cohort, reindexed consecutively and given next-state targets under the shuffled order; terminal intervals were removed. Normalisation retained the ordered training statistics. Training and selection used shuffled training and validation sequences, whereas evaluation used the original ordered cohorts. This control therefore tests shuffled-to-ordered transfer, not the isolated marginal effect of temporal adjacency, because both the history--target relation and ordering differed between training and evaluation.

For the support-alignment control, response inputs, coordinate targets and temporal order were retained. For each seed, support profiles were permuted within each cohort, reassigning alignment, decomposition, mapped/unmapped composition and exposure fraction while preserving total support activity where available. Normalisation used the unperturbed training cohort, and transformed inputs were used throughout training and evaluation.

All controls used a common causal encoder, optimisation settings, grid and fixed \(K=6\) partition; probes were fitted on training users where required, models were selected on validation, and confirmation involved no updating. They distinguish state accessibility, temporal organisation and support-alignment information but not the marginal contribution of individual multitask losses.

\subsection*{Macrostate bottleneck retention and residual information}

The predictive-state model remained frozen. Training-fitted readouts, residual
maps and diagnostic partitions were applied unchanged to validation and
confirmation. Using a uniform sample of at most 600,000 training intervals
drawn with seed 42, we compared the full hidden state, the two-coordinate
bottleneck and a residual hidden representation. Ridge regressions used
standardised predictors, an intercept and \(\alpha=1\)
\cite{Hoerl1970,Pedregosa2011}.

Full-hidden probes mapped \(\mathbf h_n^{-}\) to \(\mathbf X_n\) and \(\mathbf h_n^{+}\) to \(\mathbf X_{n+1}\). The bottleneck used the model readout as its current state and fitted next-state closure from

\[
\boldsymbol\phi(\widehat{\mathbf X}_n)
=
\left(
\widehat M_n,\widehat\Psi_n,
\widehat M_n^2,\widehat\Psi_n^2,
\widehat M_n\widehat\Psi_n
\right)^\top,
\]
which introduced no additional state dimensions. Separate maps \(g^{-}\) and
\(g^{+}\) defined

\[
\begin{aligned}
\mathbf h_{n,\mathrm{res}}^{-}
  &= \mathbf h_n^{-}
     -g^{-}[\boldsymbol\phi(\widehat{\mathbf X}_n)],\\
\mathbf h_{n,\mathrm{res}}^{+}
  &= \mathbf h_n^{+}
     -g^{+}[\boldsymbol\phi(\widehat{\mathbf X}_{n+1})].
\end{aligned}
\]

Current- and next-state probes were fitted to these residuals, with outputs clipped to \([-1,1]\). Residualisation was a regression diagnostic, not a neural intervention or orthogonal projection.

Task information was measured with standardised logistic-regression classifiers predicting whether current-bundle accuracy was at least 0.5. For each representation, we also fitted a diagnostic six-cluster mini-batch \(k\)-means partition \cite{Sculley2010}; normalised mutual information (NMI) and adjusted Rand index (ARI) measured agreement with the frozen empirical mesostates \cite{Strehl2002,HubertArabie1985}. These clusters did not define transition states.

Held-out evaluation reused coordinate, closure, learned-plane drift and transition diagnostics with training-derived standardisation, ordered centres and labels; neither \(k\)-means nor \(K\) was refitted. For Fig.~\ref{fig:event_ssl_representation}, descriptive scores were recomputed from raw held-out metrics. Correlations and cosines were mapped to \((r+1)/2\), one-step RMSE to \(1/(1+\mathrm{RMSE}/0.15)\), row-wise total variation to \(1-\mathrm{TV}\), and proportions were retained. Coordinate, closure and drift scores each averaged two components; the transition score averaged \(1-\mathrm{TV}\), transformed self-transition correlation, diagonal-dominance agreement and top-edge overlap. The descriptive composite was

\[
C_{\mathrm{struct}}
=
\frac14\left(
C_{\mathrm{coord}}+C_{\mathrm{closure}}
+C_{\mathrm{drift}}+C_{\mathrm{transition}}
\right).
\]

Retention was the ratio of each domain score to its full-hidden counterpart. Raw metrics remained primary; task and clustering metrics remained separate, and no summary entered training or model selection.

In a post hoc sensitivity analysis, regularisation-matched quadratic and low-capacity cubic tensor-spline mean closures were fitted to the fixed two-coordinate readout in training data and selected by validation coordinate mean-squared error. A continuous Gaussian closure added cross-fitted state-dependent variances and a global residual correlation; validation negative log-likelihood selected its variance regularisation without mesostate labels. Transition probabilities were obtained by Gauss--Hermite quadrature over the fixed \(K=6\) partition. We compared the learned-plane diagnostic with a matched-origin diagnostic conditioning empirical and model updates on a common current-state prediction. Model-specific 50-permutation references were descriptive. Specifications were set before confirmation; the analysis was repeated across five additional seeds.

\subsubsection*{Hidden-state geometry and probe capacity}

Using a uniform sample of at most 300,000 training intervals drawn with seed 42, we standardised pre-interval hidden activity and fitted a 64-component principal component analysis (PCA) and a two-component canonical correlation analysis (CCA) with the empirical state \cite{Pearson1901,Hotelling1936}. Correlations with the first 16 principal components measured leading-direction alignment. Ridge probes assessed current- and next-state accessibility; histogram gradient boosting predicted \(M_n,\Psi_n\) from the 64 PCA scores \cite{Friedman2001,Pedregosa2011} without reconstructing drift or transitions.

In a post hoc analysis, predictive-state, pure SSL and task-only activity were evaluated at identical sampled intervals (at most 300,000 training and 250,000 held-out states). For each model, standardisation, PCA, two-component CCA and ridge probes were fitted on the training sample. A distinct pair among the first 16 principal components was selected from training correlations to represent \(M,\Psi\) and retained for held-out evaluation. For seed 42, nonlinear gain was measured against ridge using the same PCA scores; other diagnostics were repeated across six seeds.

Within the fitted 64-component PCA subspace, eigenvalues \(\lambda_j\) gave participation ratio \((\sum_j\lambda_j)^2/\sum_j\lambda_j^2\) and effective rank \(\exp[-\sum_jp_j\log p_j]\), with \(p_j=\lambda_j/\sum_k\lambda_k\) \cite{RoyVetterli2007}. The two-nearest-neighbour (TwoNN) estimator used at most 10,000 standardised states \cite{Facco2017}. Training-fitted probes and projections were applied unchanged to held-out cohorts. CCA was a supervised diagnostic; because controls differed in training objectives and model selection, comparisons were descriptive rather than single-factor causal ablations.

\subsubsection*{Cross-model closure comparison}

\(3{,}233{,}208\) confirmation-set predictions from each model were matched one-to-one by user and bundle index, covering \(56{,}195\) users without subsampling. Current states and empirical next-state targets matched numerically; neither model was retrained or fitted to the other. Agreement was assessed intervalwise and after aggregation to the state grid and fixed \(K=6\) partition. Interval comparisons used next states and displacements; jointly supported cells supplied user-balanced drift-vector and drift-speed correlations, occupancy-weighted local cosine and field RMSE; transitions used common state assignments. A six-seed analysis compared the raw field correlation with the product of the two model--empirical field correlations and computed the partial correlation.

\subsubsection*{Sensitivity analyses}

Sensitivity analyses did not enter model selection. Coordinate variants altered memory or activity mappings and re-estimated the training field and convergence core before validation, without using confirmation data or re-estimating downstream models. Other analyses kept the state definition and models fixed while applying 1,000 positive-exponential learner reweightings, strict user-equal estimands, four grids, 50 within-learner representation permutations and seven descriptive-score definitions; confirmation comparisons used identical multipliers.

A post hoc alignment-specificity analysis compared \((M,\Psi)\) with a denominator-matched activity--idle state using identical memory and activity mappings, eligible observations, weights, grid and construction-null strata. The endpoint was fixed in validation before confirmation; no between-coordinate \(p\) value was computed, and downstream models, mesostates and residence analyses were unchanged (Supplementary Note~1 and Supplementary Table~2, panel~\textbf{c}).

A separate post hoc analysis used 2,000 paired learner-cluster bootstrap replicates for confirmation fields and fixed \(K=6\) transitions. Fixed-support intervals were standard-error limits centred on unperturbed estimates, using Fisher-\(z\) for Pearson correlations and native scales otherwise; support-reselected intervals used percentiles after reapplying the 30-transition threshold. Six-state correlations remained descriptive, with Spearman and leave-one-state-out summaries. Across 32 complementary user partitions, exact cyclic-shift fields supplied attenuation benchmarks from geometric means of Spearman--Brown-corrected split-half correlations; validation-derived activity cutpoints were unchanged in confirmation.

The 1,000 learner reweightings also re-estimated convergence thresholds and regions on fixed support while retaining the prespecified core definition. Their 2.5--97.5\% ranges quantify learner-composition sensitivity, not confidence intervals around the unperturbed estimate. Six-seed Student-\(t\) intervals quantify training-run variation; six-state correlations remain descriptive. Held-out metrics remained primary.

\subsection*{Construction-null-referenced downstream recovery}

For this post hoc analysis, the empirical field \(\mathbf b\), exact construction-null expectation \(\mathbf b_0\) and model fields \(\mathbf b_k\) used the same empirical current states, user-balanced weights, \(40\times40\) grid and fixed support. Within each opportunity-matched group, \(\mathbf b_0\) averaged all permitted non-zero cyclic shifts. The original 100-replicate ensemble remained the sole basis for the prespecified construction-null tests.

Only empirical-anchor increments, \(\widehat{\mathbf X}^{(k)}_{n+1}-\mathbf X_n\), entered this decomposition; learned-plane dynamics were analysed separately. With \(\omega_c\) denoting empirical occupancy renormalised over support \(C\),
\[
D_w^2(\mathbf f,\mathbf g)
=
\sum_{c\in C}\omega_c
\left\|\mathbf f(c)-\mathbf g(c)\right\|_2^2,
\qquad
D_w(\mathbf f,\mathbf g)=\sqrt{D_w^2(\mathbf f,\mathbf g)}.
\]
Recovery relative to the exact null was
\[
\begin{aligned}
\Delta D_k^2
&=D_w^2(\mathbf b_k,\mathbf b)-D_w^2(\mathbf b_0,\mathbf b),\\
S_k
&=1-\frac{D_w^2(\mathbf b_k,\mathbf b)}{D_w^2(\mathbf b_0,\mathbf b)},
\qquad
R_k
=\frac{D_w(\mathbf b_k,\mathbf b)}{D_w(\mathbf b_0,\mathbf b)}.
\end{aligned}
\]
Thus, \(\Delta D_k^2<0\), \(S_k>0\) and \(R_k<1\) favour the model. Secondary effect sizes compared \(\mathbf b_k-\mathbf b_0\) with \(\mathbf b-\mathbf b_0\) by vector correlation, occupancy-weighted local cosine and zero-intercept amplitude slope; coordinate-specific skills applied \(S_k\) to each axis.

An occupancy-weighted, non-negative, no-intercept rescaling \(\widetilde{\mathbf b}_0=(\alpha_M b_{0,M},\alpha_\Psi b_{0,\Psi})\) was estimated in validation and applied unchanged in confirmation. Learner-composition sensitivity used 1,000 paired positive-exponential multipliers with fixed support and null construction. The endpoint was defined after model finalisation and is reported as post hoc.

\section*{Data availability}

The EdNet-KT4 interaction logs and associated EdNet Contents metadata are publicly available from the official EdNet repository (\url{https://github.com/riiid/ednet}) under the Creative Commons Attribution--NonCommercial 4.0 International licence (\url{https://creativecommons.org/licenses/by-nc/4.0/}). Use was confined to non-commercial scholarly research. The release uses supplied user identifiers and fixed timestamp shifts for security; its collection and hierarchy are described in Ref.~\cite{Choi2020}.

\section*{Ethics statement}

This study was a secondary analysis of publicly released EdNet interaction records. No participants were recruited or contacted, no new human data were collected, and the authors did not attempt to re-identify users or link the records to external personal data. Only aggregate, non-identifying results are reported.

\section*{Code availability}
Code for all analyses is available at \url{https://github.com/MinlinWu/EdDynamics}.

\section*{Acknowledgements}

The authors thank the members of the Tianli--Beihang Joint Laboratory and the China Telecom Research Institute for their helpful discussions, and Yongqiang Luo for his support. This work was funded by Sichuan Qiming Daren Technology Co., Ltd., which also provided the computational infrastructure used in this study. ChatGPT was used to support research coding and manuscript preparation; all outputs were reviewed and verified by the authors.

\section*{Supplementary Information}

\renewcommand{\tablename}{Supplementary Table}
\renewcommand{\figurename}{Supplementary Fig.}
\setcounter{table}{0}
\setcounter{figure}{0}
\renewcommand{\theHtable}{S\arabic{table}}
\renewcommand{\theHfigure}{S\arabic{figure}}

\subsection*{Supplementary Note 1. Construction-aware validation of the empirical effective field}

The construction-matched accounting null preserved current \(M,\Psi\), denominator increments, user-balanced weights, field support and the training-defined core and shell. It reassigned the joint normalised response--exposure innovation pair within opportunity-matched strata without refitting coordinates, regions, partitions or downstream models. Matching proceeded from finer within-user strata to across-user strata, with global groups used only when necessary; continuous cutpoints were estimated from training data.

Within-user strata matched response, support, idle and observation-gap opportunity. Across-user strata additionally matched fixed bins of response and exposure mass, support share, idle share and sequence length. Validation and confirmation each used 100 permutations and were analysed separately.

The observed field departed from matched accounting in both held-out cohorts (Supplementary Table~\ref{tab:supp_construction_null}). Validation supported the full-field and negative-divergence departures; confirmation replicated both and additionally supported shell inward flow, whereas core slowing did not differ from matched accounting in either cohort. After null subtraction, excess fields agreed across 917 cells (vector \(r=0.7164\), speed \(r=0.7362\), occupancy-weighted local cosine \(0.9604\)). Normalised-memory accounting therefore contributes generic relaxation but does not determine the reproducible state-conditioned field.

A post hoc alignment-specificity control retained \(M\) but replaced \(\Psi\) with the activity--idle balance \(\Phi=(H^++H^-+H^0-I)/(H^++H^-+H^0+I)\), which does not use content--demand alignment and retains the same 10-day exposure memory and active-plus-idle denominator. Activity and idle mappings, eligible observations, user weights, grid thresholds and construction-null opportunity strata were unchanged; each coordinate system retained its own training-defined core, and the validation endpoint was fixed before the confirmation cohort was evaluated. Both states reproduced raw fields and departed from their matched nulls, but the alignment-based state showed approximately threefold larger field-RMS-normalised departure in validation (\(0.3808\) versus \(0.1300\)) and confirmation (\(0.3785\) versus \(0.1236\)). This single-comparator result identifies greater null-referenced prominence, not causal necessity or coordinate optimality: \(\Phi\) retains event-type semantics, is strongly boundary-concentrated and provides no between-coordinate \(p\) value (Supplementary Table~\ref{tab:supp_empirical_sensitivity}, panel~\textbf{c}).

A post hoc whole-user analysis rebuilt exact cyclic-shift expectations in 32 complementary partitions using activity cutpoints estimated in validation and applied unchanged in confirmation. Permutation-based and exact point estimates differed by at most \(0.0041\). Across 29 valid partitions, median vector and speed agreement-to-benchmark ratios were \(1.023\) and \(0.976\), with 2.5--97.5\% ranges \(0.963\)--\(1.067\) and \(0.939\)--\(1.017\). These are attenuation benchmarks, not latent-truth estimates; prespecified null inference remains permutation-based.

All 1,000 learner reweightings recovered dynamically qualified regions in both cohorts. The selection-frequency overlap coefficient was \(0.785\), while the thresholded-region Jaccard varied (median \(0.456\), 2.5--97.5\% \(0.167\)--\(0.605\)) and median centre separation was \(0.110\). Thus, the region was persistently detected and spatially aligned despite threshold-sensitive boundaries; the prespecified core was unchanged.

The empirical field was also stable to the declared coordinate mappings and observation-gap restrictions (Supplementary Table~\ref{tab:supp_empirical_sensitivity}). All four memory or activity variants produced training cores satisfying the prespecified dynamical criteria and retained contractive, inward and centrally slowed validation flow. The slow-activity mapping displaced the training core most strongly, but its validation field still correlated \(0.9622\) with the primary field. Restricting to observed successors or excluding terminal and \(>7\)-day gaps left field direction and speed almost unchanged in both held-out cohorts.

\begin{table*}[t]
\centering
\caption{\textbf{Construction-matched accounting-null tests and matching
quality.}
\textbf{a}, Prespecified validation tests and held-out confirmation replication without updating. The full-field distance is tested separately from the three basin metrics. \(q_{\mathrm{BH}}\) denotes the Benjamini--Hochberg-adjusted value for negative divergence, shell inward flow and core slowing. With 100 permutations, the minimum \(+1\)-corrected \(p_{\mathrm{MC}}\) is \(0.0099\). \textbf{b}, Fractions assigned by the within-user, across-user, global-opportunity and weak-fallback matching tiers; the joint response--exposure innovation pair moved without fixed points. \textbf{c}, Opportunity composition and numerical identity. Reconstruction error is the maximum across next-state coordinates; the final column checks the frozen empirical drift and occupancy against their reference values.}
\label{tab:supp_construction_null}
\begingroup
\scriptsize
\setlength{\tabcolsep}{3.6pt}
\renewcommand{\arraystretch}{1.10}

\textbf{a. Prespecified field and basin tests}\par\vspace{2pt}

\resizebox{\textwidth}{!}{%
\begin{tabular}{llrrrrr}
\toprule
Split & Metric & Observed & Null mean & Null 2.5--97.5\% &
\(p_{\mathrm{MC}}\) & \(q_{\mathrm{BH}}\) \\
\midrule
\multirow{4}{*}{validation}
& Full-field distance & 0.09097 & 0.00934 & [0.00891, 0.00991] & 0.0099 & -- \\
& Negative-divergence occupancy & 0.7762 & 0.7378 & [0.7184, 0.7596] & 0.0099 & 0.0297 \\
& Frozen-shell inward fraction & 0.6693 & 0.6625 & [0.6541, 0.6695] & 0.0396 & 0.0594 \\
& Core-to-shell speed ratio & 0.5226 & 0.4921 & [0.4777, 0.5050] & 1.0000 & 1.0000 \\
\addlinespace
\multirow{4}{*}{confirmation}
& Full-field distance & 0.09074 & 0.00916 & [0.00878, 0.00956] & 0.0099 & -- \\
& Negative-divergence occupancy & 0.8103 & 0.7382 & [0.7157, 0.7567] & 0.0099 & 0.0149 \\
& Frozen-shell inward fraction & 0.6769 & 0.6642 & [0.6567, 0.6714] & 0.0099 & 0.0149 \\
& Core-to-shell speed ratio & 0.5047 & 0.4919 & [0.4807, 0.5049] & 0.9703 & 0.9703 \\
\bottomrule
\end{tabular}%
}

\vspace{5pt}
\textbf{b. Opportunity-matching audit}\par\vspace{2pt}

\resizebox{\textwidth}{!}{%
\begin{tabular}{lrrrrrrr}
\toprule
Split & Transitions & Replicates & Within user & Across user &
Global opportunity & Weak-fallback tier & Fixed points \\
\midrule
validation &
\(3{,}328{,}409\) & 100 & 0.995183 & 0.004814 &
\(2.704\times10^{-6}\) & \(9.013\times10^{-7}\) & 0 \\
confirmation &
\(3{,}233{,}208\) & 100 & 0.995002 & 0.004995 &
\(3.712\times10^{-6}\) & 0 & 0 \\
\bottomrule
\end{tabular}%
}
\vspace{5pt}
\textbf{c. Opportunity composition and numerical identity}\par\vspace{2pt}

\resizebox{\textwidth}{!}{%
\begin{tabular}{lrrrrrr}
\toprule
Split &
\shortstack{Randomisable\\fraction} &
\shortstack{Zero-innovation\\rows} &
\shortstack{Support-present\\fraction} &
\shortstack{Idle-present\\fraction} &
\shortstack{Maximum next-state\\reconstruction error} &
\shortstack{Maximum frozen field/\\occupancy difference} \\
\midrule
validation &
1.000000 &
0 &
0.874112 &
0.999978 &
\(2.76\times10^{-14}\) &
0 \\
confirmation &
1.000000 &
0 &
0.869599 &
0.999976 &
\(2.14\times10^{-14}\) &
0 \\
\bottomrule
\end{tabular}%
}
\endgroup
\end{table*}

\begin{table*}[t]
\centering
\caption{\textbf{Coordinate-construction, observation-gap and alignment-specificity controls.} \textbf{a}, Coordinate variants altered only memory or activity mappings, re-estimated the training field and core, and were evaluated on validation without using confirmation data or repeating downstream analyses. \(r_{\mathrm{primary}}\) compares each variant with the primary validation field; \(c_{\mathrm{rep}}\) is the training--validation mean local cosine. \textbf{b}, Observation restrictions relative to the field estimated from all eligible transitions in the same split; the last three columns show restricted-minus-all-row changes. Coordinates, thresholds, core, shell and the \(K=6\) partition were fixed. \textbf{c}, Post hoc comparison of the alignment-based state with one denominator-matched activity--idle comparator. \(Q_X\) is the occupancy-weighted distance from each coordinate's own construction-null mean divided by its held-out supported-field RMS. \(\sigma_2\) and \(f_{\mathrm{bdry}}\) are the second-axis standard deviation and boundary fraction \(\Pr(|x_2|\geq0.95)\); \(\widetilde f_{\Delta Z_2}\) and \(f_{\mathrm{var}}\) audit second-axis innovation change and matched-group variation. Between-coordinate differences are descriptive; \(p_{\mathrm{MC}}\) tests each coordinate against its own null, and separately selected core/shell metrics are not winner criteria.}
\label{tab:supp_empirical_sensitivity}
\begingroup
\scriptsize
\setlength{\tabcolsep}{3.2pt}
\renewcommand{\arraystretch}{1.10}

\textbf{a. Declared memory and activity mappings}\par\vspace{2pt}

\resizebox{\textwidth}{!}{%
\begin{tabular}{lllrrrrrl}
\toprule
Setting & \(\tau_R/\tau_A\) (d) & Activity/idle mapping &
\(r_{\mathrm{primary}}\) & \(c_{\mathrm{rep}}\) &
\(f_{\mathrm{neg}}\) & \(f_{\mathrm{in}}\) & \(R_{\mathrm{cs}}\) &
Training-core centre \\
\midrule
Primary & 10/10 & 3/2.5/4 min; 1 d &
1.0000 & 0.9657 & 0.7762 & 0.6693 & 0.5226 &
\((-0.0197,-0.5820)\) \\
Memory 5 d & 5/5 & 3/2.5/4 min; 1 d &
0.9779 & 0.9697 & 0.7863 & 0.8491 & 0.4227 &
\((-0.1159,-0.6609)\) \\
Memory 20 d & 20/20 & 3/2.5/4 min; 1 d &
0.9805 & 0.9679 & 0.8020 & 0.6903 & 0.5420 &
\((-0.0141,-0.5760)\) \\
Activity fast & 10/10 & 2/2/3 min; 0.5 d &
0.9797 & 0.9679 & 0.7652 & 0.6630 & 0.5584 &
\((-0.0252,-0.5656)\) \\
Activity slow & 10/10 & 4/4/6 min; 2 d &
0.9622 & 0.9638 & 0.8310 & 0.9569 & 0.6327 &
\((-0.3422,-0.8111)\) \\
\bottomrule
\end{tabular}%
}

\vspace{5pt}
\textbf{b. Observation-gap restrictions}\par\vspace{2pt}

\resizebox{\textwidth}{!}{%
\begin{tabular}{llrrrrrrrr}
\toprule
Split & Restriction & Intervals & Cells &
\(r_{\mathbf b}\) & \(\overline c_{\mathrm{local}}\) &
\(r_{\lVert\mathbf b\rVert}\) &
\(\Delta f_{\mathrm{neg}}\) & \(\Delta f_{\mathrm{in}}\) &
\(\Delta R_{\mathrm{cs}}\) \\
\midrule
validation & Observed successors only &
\(3{,}328{,}409\) & 940 & 0.9996 & 0.9998 & 0.9993 &
+0.0047 & \(-0.0020\) & +0.0092 \\
validation & Exclude terminal or \(>7\)-day gaps &
\(3{,}314{,}344\) & 938 & 0.9978 & 0.9989 & 0.9966 &
+0.0033 & \(-0.0073\) & +0.0125 \\
confirmation & Observed successors only &
\(3{,}233{,}208\) & 954 & 0.9996 & 0.9999 & 0.9995 &
+0.0062 & \(-0.0001\) & +0.0088 \\
confirmation & Exclude terminal or \(>7\)-day gaps &
\(3{,}218{,}913\) & 953 & 0.9972 & 0.9988 & 0.9953 &
+0.0030 & \(-0.0019\) & +0.0104 \\
\bottomrule
\end{tabular}%
}

\vspace{5pt} 
\textbf{c. Post hoc alignment-specificity control}\par\vspace{2pt} \resizebox{\textwidth}{!}{%
\begin{tabular}{llrrrrrrr} 
\toprule 
Split & State & 
\(r_{\mathbf b}\) & 
\(\overline c_{\mathrm{local}}\) & 
\shortstack{Supported cells/\\occupancy mass/Jaccard} & 
\(Q_X\) & 
\(p_{\mathrm{MC}}\) & 
\(\sigma_2/f_{\mathrm{bdry}}\) & 
\(\widetilde f_{\Delta Z_2}/f_{\mathrm{var}}\) \\ 
\midrule 
\multirow{2}{*}{validation} 
& Alignment-based \((M,\Psi)\) & 
0.9424 & 0.9748 & 940/0.9963/0.8735 & 
0.3808 & 0.0099 & 0.2035/0.0005 & --/-- \\ 
& Activity--idle \((M,\Phi)\) & 
0.9153 & 0.9698 & 296/0.9984/0.7851 & 
0.1300 & 0.0099 & 0.0384/0.8151 & 0.9983/1.0000 \\ 
\addlinespace 
\multirow{2}{*}{confirmation} 
& Alignment-based \((M,\Psi)\) & 
0.9430 & 0.9829 & 954/0.9961/0.8848 & 
0.3785 & 0.0099 & 0.2071/0.0002 & --/-- \\ 
& Activity--idle \((M,\Phi)\) & 
0.9194 & 0.9866 & 296/0.9982/0.7851 & 
0.1236 & 0.0099 & 0.0388/0.8032 & 0.9983/1.0000 \\ 
\bottomrule 
\end{tabular}%
}
\endgroup
\end{table*}

\subsection*{Supplementary Note 2. Mesostate kinetics and construction-aware robustness}

The continuous field was estimated without clustering. The fixed \(K=6\) partition supplied only an operational kinetic coarse-graining, with centres fitted on training data and applied unchanged. All six validation rows satisfied \(P_{ii}=\max_j P_{ij}\), but their persistence strengths and residence profiles were heterogeneous. Reliable-horizon restricted mean residence exceeded the state-matched geometric reference in all six mesostates, whereas positive survival excess at the prespecified 10-step horizon was confined to \(S_0,S_4,S_5\) (Supplementary Table~\ref{tab:supp_empirical_kinetics}).

Robustness analyses sharpened the state-heterogeneous interpretation (Supplementary Table~\ref{tab:supp_kinetic_robustness}). The recursive surrogate preserved the fixed opportunity-matching scheme, denominators, segments and observation structure while propagating coherent state sequences. The post hoc all-state endpoint did not support a uniform 10-step lift, whereas the post hoc family-wise maxT test identified \(S_0,S_4,S_5\); the learner-cluster bootstrap yielded reliable-horizon RMST lift lower bounds above one in all six mesostates and retained positive 10-step excess in the same three. RMST lift also remained above one in every mesostate for \(K=4\)--8. These results separate broad reliable-horizon persistence from fixed-horizon excess concentrated in particular mesostates.

Here, metastable-like refers to reproducible, population-level, state-heterogeneous basin and residence kinetics under the stated Kaplan--Meier censoring assumption, not spectral metastability, autonomous Markov closure or causal platform effects.

\begin{table*}[t]
\centering
\caption{\textbf{Fixed six-state kinetic coarse-graining.} Training-defined centres and validation kinetics under the fixed \(K=6\) partition. Occupancy is user-balanced, and residence uses censor-aware Kaplan--Meier estimates. \(\tau_i\) is the largest reliable horizon; RMST lift is the observed-to-geometric restricted-mean ratio at \(\tau_i\), so statewise lift magnitudes are not directly comparable. \(D_{10}=\widehat S_i(10)-P_{ii}^{9}\). The final column gives the Benjamini--Hochberg-adjusted one-sided \(q\) value from Greenwood-variance inference at 10 steps.}
\label{tab:supp_empirical_kinetics}
\begingroup
\scriptsize
\setlength{\tabcolsep}{3.5pt}
\renewcommand{\arraystretch}{1.12}
\resizebox{\textwidth}{!}{%
\begin{tabular}{lrrrrrrrr}
\toprule
State & Centre \((M,\Psi)\) & Occupancy & \(P_{ii}\) &
Episodes / censored fraction & \(\tau_i\) / RMST lift &
\(n_{\mathrm{risk}}(10)\) & \(D_{10}\) & Greenwood \(q\) \\
\midrule
\(S_0\) & \((-0.8830,\,0.0087)\) & 0.1010 & 0.3623 &
\(36{,}704/0.0510\) & \(74/1.1503\) & 318 & 0.0183 &
\(7.76\times10^{-92}\) \\
\(S_1\) & \((-0.7098,\,-0.6699)\) & 0.2068 & 0.8846 &
\(45{,}062/0.3293\) & \(553/2.0364\) & \(4{,}986\) & \(-0.0425\) &
1.0000 \\
\(S_2\) & \((-0.2868,\,-0.4409)\) & 0.1560 & 0.9069 &
\(57{,}617/0.1846\) & \(1{,}189/1.8592\) & \(6{,}814\) & \(-0.2228\) &
1.0000 \\
\(S_3\) & \((0.1026,\,-0.7116)\) & 0.2719 & 0.9872 &
\(51{,}820/0.4235\) & \(4{,}302/6.4733\) & \(14{,}961\) & \(-0.4026\) &
1.0000 \\
\(S_4\) & \((0.1737,\,-0.1317)\) & 0.1055 & 0.7330 &
\(32{,}010/0.1720\) & \(138/1.2776\) & \(1{,}362\) & 0.0240 &
\(1.95\times10^{-33}\) \\
\(S_5\) & \((0.9661,\,-0.0017)\) & 0.1587 & 0.6238 &
\(32{,}796/0.0868\) & \(35/1.1272\) & 454 & 0.0311 &
\(1.06\times10^{-107}\) \\
\bottomrule
\end{tabular}%
}
\endgroup
\end{table*}

\begin{table*}[t]
\centering
\caption{\textbf{Construction-aware and learner-cluster robustness of mesostate kinetics.} \textbf{a}, The post hoc recursive aggregate endpoint retains the prespecified 10-step horizon and tests the all-state mean log RMST lift; the post hoc studentised maxT endpoint tests positive \(D_{10}\) in any fixed mesostate. Both use the same 100 surrogates and the \(+1\) correction. \textbf{b}, Statewise maxT and 1,000-replicate learner-cluster results; \(q_{\mathrm{cl}}\) uses a one-sided normal approximation with Benjamini--Hochberg adjustment, and \(\Pr(\mathrm{diag})\) is the bootstrap frequency of \(P_{ii}=\max_j P_{ij}\). \(p_{\mathrm{FWER}}\) denotes the maxT \(p\) value adjusted for the family-wise error rate. \textbf{c}, Partition-resolution checks; \(K=6\) is the primary partition, and states are not matched across other values of \(K\).}
\label{tab:supp_kinetic_robustness}
\begingroup
\scriptsize
\setlength{\tabcolsep}{3.2pt}
\renewcommand{\arraystretch}{1.10}

\textbf{a. Recursive construction- and denominator-inertia-matched surrogate}\par\vspace{2pt}

\resizebox{\textwidth}{!}{%
\begin{tabular}{lrrrrl}
\toprule
Endpoint & Observed & Null median & Null 2.5--97.5\% &
\(p_{\mathrm{MC}}\) & Status \\
\midrule
Mean log RMST lift through 10 steps & \(-0.2280\) & \(-0.1515\) &
\([-0.1535,-0.1489]\) & 1.0000 & Post hoc aggregate robustness endpoint \\
Mean self-transition probability & 0.7496 & 0.6728 &
[0.6713, 0.6740] & -- & Descriptive \\
Rows with \(P_{ii}=\max_j P_{ij}\) & 6 & 5 & [5, 5] & -- & Descriptive \\
Statewise studentised maxT at 10 steps & 25.2098 & 1.2503 &
[0.1869, 2.9769] & 0.0099 & \(S_0,S_4,S_5\) family-wise significant \\
\bottomrule
\end{tabular}%
}

\vspace{5pt}
\textbf{b. Primary \(K=6\) statewise inference}\par\vspace{2pt}

\resizebox{\textwidth}{!}{%
\begin{tabular}{lrrrrrrr}
\toprule
State & \(D_{10}\) & Recursive-null 2.5--97.5\% range &
\(p_{\mathrm{FWER}}\) & Learner-cluster \(D_{10}\) 95\% CI &
\(q_{\mathrm{cl}}\) & RMST lift 95\% CI & \(\Pr(\mathrm{diag})\) \\
\midrule
\(S_0\) & 0.0183 & [0.0055, 0.0073] & 0.0099 &
[0.0163, 0.0203] & \(1.24\times10^{-71}\) &
[1.0691, 1.2059] & 0.8670 \\
\(S_1\) & \(-0.0425\) & [0.0834, 0.0907] & 1.0000 &
\([-0.0641, -0.0216]\) & 1.0000 &
[1.9490, 2.1351] & 1.0000 \\
\(S_2\) & \(-0.2228\) & \([-0.0302, -0.0225]\) & 1.0000 &
\([-0.2540, -0.1953]\) & 1.0000 &
[1.7722, 1.9472] & 1.0000 \\
\(S_3\) & \(-0.4026\) & \([-0.4080, -0.4023]\) & 0.2178 &
\([-0.4097, -0.3957]\) & 1.0000 &
[6.0880, 6.8551] & 1.0000 \\
\(S_4\) & 0.0240 & \([-0.0119, -0.0015]\) & 0.0099 &
[0.0169, 0.0298] & \(8.35\times10^{-13}\) &
[1.2504, 1.3015] & 1.0000 \\
\(S_5\) & 0.0311 & [0.0262, 0.0285] & 0.0099 &
[0.0288, 0.0332] & \(3.19\times10^{-172}\) &
[1.1157, 1.1390] & 1.0000 \\
\bottomrule
\end{tabular}%
}

\vspace{5pt}
\textbf{c. Bounded partition-resolution sensitivity}\par\vspace{2pt}

\resizebox{\textwidth}{!}{%
\begin{tabular}{ccccccccc}
\toprule
\(K\) & Role & Minimum occupancy & Diagonal rows &
Mean \(P_{ii}\) (range) & RMST lift \(>1\) & Positive \(D_{10}\) &
Train--validation row TV & Mean \(\lvert\Delta\log L_{\mathrm{RMST}}\rvert\) \\
\midrule
4 & Sensitivity & 0.1679 & 4/4 & 0.8228 (0.6676--0.9855) &
4/4 & 2/4 & 0.0041 & 0.0634 \\
5 & Sensitivity & 0.1031 & 4/5 & 0.7599 (0.3888--0.9870) &
5/5 & 2/5 & 0.0061 & 0.0772 \\
6 & Primary & 0.1010 & 6/6 & 0.7496 (0.3623--0.9872) &
6/6 & 3/6 & 0.0062 & 0.0571 \\
7 & Sensitivity & 0.0917 & 7/7 & 0.7439 (0.3413--0.9736) &
7/7 & 4/7 & 0.0059 & 0.0789 \\
8 & Sensitivity & 0.0674 & 7/8 & 0.7233 (0.2921--0.9802) &
8/8 & 4/8 & 0.0070 & 0.0706 \\
\bottomrule
\end{tabular}%
}
\endgroup
\end{table*}

\subsection*{Supplementary Note 3. Mechanism-family selection and post-selection kinetics}

Mechanism selection was restricted to the declared family hierarchy, finite search domain and composite scoring rule. Under the prespecified weights, the seven-term reference had the lowest bootstrap mean score, while the four-term offset dual-channel family was the simplest model meeting both eligibility criteria. No family with three or fewer coefficients qualified, and selection was unchanged for practical-equivalence margins \(0.010\)--\(0.030\) (Supplementary Table~\ref{tab:supp_mechanism_family}).

A post hoc analysis using only training and validation data left the prespecified confirmation evaluation unchanged. Repeating the complete selection procedure with equal component weights again selected offset dual-channel (\(k=4\)); the paired difference from the seven-term best was \(0.00616\) (95\% interval, \(0.00497\)--\(0.00747\)). No family with \(k\leq4\) Pareto-dominated it across the five losses and parameter count in any of 300 paired-user bootstraps. Four-term parsimony therefore remained stable under equal weighting but is bounded by the declared objective, hierarchy and search domain.

The four family-varying substantive coefficients were frozen at
\[
(\theta_0,\theta_M,\phi_0,\delta_S)=(-0.17,\,0.64,\,-1.35,\,6),
\]
with structural zeros
\(\theta_{\Psi}=\theta_{M\Psi}=\phi_{\Psi}=0\). Shared scales
\((\lambda_R,\lambda_A,\lambda_I)=(0.46,1.10,0.85)\) and
training-estimated accounting quantities
\((\eta,\tau_R,\tau_A,r,\gamma_R,\gamma_A)
=(20,10,10,0.3865,1,0.9537)\)
remained part of the frozen update specification but were not additional candidate mechanism terms. Confirmation involved evaluation only, with no reconsideration of candidate families, parameter update, calibration change, region redefinition or partition refit.

Transition and residence diagnostics were excluded from selection. Five of six confirmation states retained the same dominant destination, and self-transition probabilities preserved the empirical persistence ordering (\(r=0.9870\)); the sole dominant-edge deviation was \(S_0\) (Supplementary Table~\ref{tab:supp_mechanism_kinetics}). Transition-implied residence preserved this ordering while differing in absolute tail scale, exposing long-horizon structure beyond the one-step closure. These quantities derive from the one-step transition matrices, not autonomous residence trajectories.

\begin{table*}[t]
\centering
\caption{\textbf{Complete mechanism-family comparison.} Primary-structure scores are paired validation-user bootstrap means from 300 resamples; lower is better. \(k\) counts family-varying substantive coefficients. \(\Delta_{\mathrm{best}}\) is the paired score difference from the best-scoring family in the bootstrap comparison. ``1-SE'' denotes membership in the one-standard-error set, and ``PE'' denotes practical equivalence at margin \(0.02\); both were required for selection. The last two rows are additional direct one-term deletion refits; the other two direct deletions coincide with the dual-channel and response-offset cores.}
\label{tab:supp_mechanism_family}
\begingroup
\scriptsize
\setlength{\tabcolsep}{2.8pt}
\renewcommand{\arraystretch}{1.08}
\resizebox{\textwidth}{!}{%
\begin{tabular}{lclccccc}
\toprule
Family & \(k\) & Active substantive terms &
Mean score [95\% bootstrap interval] & \(\Delta_{\mathrm{best}}\) [95\%] &
1-SE & PE & Selected \\
\midrule
State persistence & 0 & -- &
0.8080 [0.8078, 0.8081] & 0.5489 [0.5326, 0.5633] &
No & No & No \\
Alignment-only signed mechanism & 1 & \(\phi_0\) &
0.4431 [0.4278, 0.4601] & 0.1840 [0.1695, 0.1968] &
No & No & No \\
Response-only signed mechanism & 2 & \(\theta_0,\theta_M\) &
0.6296 [0.6186, 0.6371] & 0.3705 [0.3565, 0.3851] &
No & No & No \\
Two-coordinate core & 2 & \(\theta_M,\phi_0\) &
0.2898 [0.2761, 0.3061] & 0.0307 [0.0247, 0.0365] &
No & No & No \\
Response-offset core & 3 & \(\theta_0,\theta_M,\phi_0\) &
0.2824 [0.2681, 0.2989] & 0.0233 [0.0184, 0.0282] &
No & No & No \\
Dual-channel core & 3 & \(\theta_M,\phi_0,\delta_S\) &
0.2715 [0.2575, 0.2880] & 0.0125 [0.0092, 0.0161] &
No & Yes & No \\
Offset dual-channel & 4 & \(\theta_0,\theta_M,\phi_0,\delta_S\) &
0.2642 [0.2496, 0.2817] & 0.0051 [0.0032, 0.0071] &
Yes & Yes & Yes \\
Dual-channel with linear coupling & 4 &
\(\theta_M,\theta_\Psi,\phi_0,\delta_S\) &
0.2716 [0.2576, 0.2893] & 0.0126 [0.0094, 0.0159] &
No & Yes & No \\
Dual-channel with interaction & 4 &
\(\theta_M,\theta_{M\Psi},\phi_0,\delta_S\) &
0.2714 [0.2577, 0.2876] & 0.0124 [0.0090, 0.0159] &
No & Yes & No \\
Full seven-term reference & 7 &
\(\theta_0,\theta_M,\theta_\Psi,\theta_{M\Psi},\phi_0,\delta_S,\phi_\Psi\) &
0.2590 [0.2446, 0.2754] & 0.0000 [0.0000, 0.0000] &
Yes & Yes & No \\
\addlinespace
Offset dual-channel without \(\theta_M\) & 3 &
\(\theta_0,\phi_0,\delta_S\) &
0.3366 [0.3224, 0.3552] & 0.0776 [0.0713, 0.0843] &
No & No & No \\
Offset dual-channel without \(\phi_0\) & 3 &
\(\theta_0,\theta_M,\delta_S\) &
0.6268 [0.6195, 0.6327] & 0.3677 [0.3525, 0.3821] &
No & No & No \\
\bottomrule
\end{tabular}%
}
\endgroup
\end{table*}

\begin{table*}[t]
\centering
\caption{\textbf{Statewise post-selection mechanism kinetics.}
Held-out confirmation results under the fixed empirical \(K=6\) partition.
\(\Delta P_{ii}=P_{ii}^{\mathrm{mech}}-P_{ii}^{\mathrm{emp}}\), and row TV is
the total-variation distance between transition rows. The top destination
includes self-transition. The final column is the ratio of
transition-implied geometric mean residence references
\(1/(1-P_{ii})\); it is not an empirical Kaplan--Meier residence ratio.}
\label{tab:supp_mechanism_kinetics}
\begingroup
\scriptsize
\setlength{\tabcolsep}{4.0pt}
\renewcommand{\arraystretch}{1.12}
\resizebox{\textwidth}{!}{%
\begin{tabular}{lrrrrccc}
\toprule
State & \(P_{ii}^{\mathrm{emp}}\) & \(P_{ii}^{\mathrm{mech}}\) &
\(\Delta P_{ii}\) & Row TV &
Empirical/mechanism top destination & Match &
\(\bar L_{\mathrm{mech}}^{\mathrm{geo}}/
 \bar L_{\mathrm{emp}}^{\mathrm{geo}}\) \\
\midrule
\(S_0\) & 0.4050 & 0.3620 & \(-0.0429\) & 0.2775 & \(S_0/S_2\) & No & 0.9327 \\
\(S_1\) & 0.8881 & 0.9592 & +0.0711 & 0.0711 & \(S_1/S_1\) & Yes & 2.7416 \\
\(S_2\) & 0.9109 & 0.9665 & +0.0556 & 0.0556 & \(S_2/S_2\) & Yes & 2.6604 \\
\(S_3\) & 0.9864 & 0.9947 & +0.0084 & 0.0084 & \(S_3/S_3\) & Yes & 2.5879 \\
\(S_4\) & 0.7634 & 0.8103 & +0.0469 & 0.0469 & \(S_4/S_4\) & Yes & 1.2474 \\
\(S_5\) & 0.6267 & 0.5508 & \(-0.0760\) & 0.1532 & \(S_5/S_5\) & Yes & 0.8309 \\
\bottomrule
\end{tabular}%
}
\endgroup
\end{table*}

\subsection*{Supplementary Note 4. Event-SSL seed robustness and hidden-state organisation}

The complete training and evaluation procedure was repeated with five additional seeds (\(2026,666,606,37,4669\)) alongside seed 42. All runs used the same prespecified state and training-defined \(K=6\) partition. Coordinate, one-step, landscape, field and transition recovery remained stable (Supplementary Table~\ref{tab:supp_event_ssl_seeds}). Learned-plane field correlation was positive in all six runs; shuffled-to-ordered transfer reversed its sign in all six, and support-alignment randomisation reduced inward transport in all six. Full Event-SSL exceeded task-only in field correlation for every seed and pure SSL for five of six.

The effective state formed a prominent but non-exhaustive component of the recurrent representation. Across seeds, the two-coordinate bottleneck retained nearly all coordinate and descriptive transition scores, about four-fifths of closure and drift scores, and roughly \(90\%\) of overall macrostructure. Linear projections slightly exceeded the trained readout, whereas residualisation depleted overall macrostructure while retaining task information.

For seed 42, local cosine and coarse transition topology exceeded within-user permutation floors, whereas global field correlation and statewise persistence did not. Alternative state-only closures improved matched-origin drift and transitions across seeds but transferred incompletely to the primary learned-plane gauge, indicating distinct contributions from closure form and current-state readout mismatch (Supplementary Table~\ref{tab:supp_representation_diagnostics}, panels~\textbf{a--d}). For the Gaussian closure, quadrature converged at order 15 and confirmation boundary-censoring mass was \(0.77\%\); no permutation \(p\) value was calculated.

A separate post hoc analysis compared the fixed predictive-state, pure-SSL and task-only representations on the same samples. Both controls retained two canonical macrostate directions (mean second canonical correlations across six seeds, \(0.817\) and \(0.810\), versus \(0.900\) for full Event-SSL), but their macrostate access was less concentrated in leading principal components and their seed-42 feature-matched nonlinear gains were larger. All three representations remained higher-dimensional by spectral diagnostics (participation ratios \(11.2\)--\(13.9\)). Encoders trained without state or closure targets therefore retained two linearly decodable axes, while the full model exhibited a more explicit linear component (Supplementary Table~\ref{tab:supp_representation_diagnostics}, panel~\textbf{e}). These geometry diagnostics are descriptive; raw held-out metrics remain primary, and objective differences are not single-factor causal effects.

\begin{table*}[t]
\centering
\caption{\textbf{Random-seed robustness of Event-SSL, controls and
representation organisation.} Seed 42 supplied the main-text estimates; the complete training and evaluation procedure was repeated with five additional seeds. Values are means \(\pm\) sample standard deviations (s.d.), 95\% Student-\(t\) intervals and observed ranges across six runs. The intervals describe variation across training seeds, not learner-population uncertainty. Paired control rows report full Event-SSL minus the corresponding control, except for the shuffled-to-ordered transfer row, which reports the control itself.}
\label{tab:supp_event_ssl_seeds}
\begingroup
\scriptsize
\setlength{\tabcolsep}{2.5pt}
\renewcommand{\arraystretch}{1.08}
\begin{tabular}{@{}p{0.35\textwidth}cccc@{}}
\toprule
Quantity & Mean \(\pm\) s.d. & 95\% seed-\(t\) interval &
Observed range & Consistency \\
\midrule
\multicolumn{5}{@{}l}{\textit{Predictive-state confirmation recovery}} \\
Coordinate correlation, \(M\) &
\(0.8842\pm0.0014\) & [0.8827, 0.8857] & [0.8824, 0.8864] & -- \\
Coordinate correlation, \(\Psi\) &
\(0.9906\pm0.0010\) & [0.9895, 0.9917] & [0.9888, 0.9920] & -- \\
One-step RMSE, \(M\) &
\(0.1097\pm0.0015\) & [0.1082, 0.1113] & [0.1076, 0.1115] & -- \\
One-step RMSE, \(\Psi\) &
\(0.0350\pm0.0014\) & [0.0335, 0.0364] & [0.0334, 0.0368] & -- \\
Next-state occupancy JS &
\(0.1982\pm0.0052\) & [0.1927, 0.2036] & [0.1912, 0.2038] & -- \\
Empirical-anchor drift correlation &
\(0.8845\pm0.0068\) & [0.8774, 0.8917] & [0.8778, 0.8969] & -- \\
Learned-plane drift correlation &
\(0.6829\pm0.0143\) & [0.6678, 0.6979] & [0.6630, 0.6952] & \(6/6>0\) \\
Learned-plane local cosine &
\(0.8505\pm0.0122\) & [0.8378, 0.8633] & [0.8285, 0.8629] & -- \\
Transition mean row TV &
\(0.1058\pm0.0090\) & [0.0964, 0.1153] & [0.0942, 0.1180] & -- \\
Statewise self-transition correlation &
\(0.7032\pm0.0764\) & [0.6230, 0.7833] & [0.6263, 0.8462] & \(6/6>0\) \\
\addlinespace
\multicolumn{5}{@{}l}{\textit{Same-seed control contrasts}} \\
Full Event-SSL minus pure SSL learned-plane drift correlation &
\(0.1113\pm0.0904\) & [0.0164, 0.2062] & \([-0.0065, 0.2358]\) & \(5/6>0\) \\
Full Event-SSL minus task-only learned-plane drift correlation &
\(0.0767\pm0.0363\) & [0.0386, 0.1147] & [0.0215, 0.1255] & \(6/6>0\) \\
Shuffled-to-ordered transfer learned-plane drift correlation &
\(-0.4338\pm0.0413\) & \([-0.4771, -0.3904]\) &
\([-0.4932, -0.3870]\) & \(6/6<0\) \\
Full Event-SSL minus support-alignment-randomised inward fraction &
\(0.1245\pm0.0943\) & [0.0256, 0.2235] & [0.0293, 0.2298] & \(6/6>0\) \\
\addlinespace
\multicolumn{5}{@{}l}{\textit{Representation organisation}} \\
CCA correlation 1 &
\(0.9912\pm0.0004\) & [0.9908, 0.9916] & [0.9907, 0.9917] & -- \\
CCA correlation 2 &
\(0.9010\pm0.0015\) & [0.8994, 0.9026] & [0.8995, 0.9037] & -- \\
Bottleneck overall retention &
\(0.9037\pm0.0186\) & [0.8842, 0.9232] & [0.8765, 0.9285] & -- \\
Bottleneck coordinate retention &
\(0.9954\pm0.0005\) & [0.9949, 0.9959] & [0.9946, 0.9960] & -- \\
Bottleneck closure retention &
\(0.7989\pm0.0026\) & [0.7962, 0.8017] & [0.7962, 0.8026] & -- \\
Bottleneck drift retention &
\(0.7958\pm0.0698\) & [0.7226, 0.8690] & [0.6961, 0.8849] & -- \\
Bottleneck transition retention &
\(0.9991\pm0.0037\) & [0.9952, 1.0031] & [0.9942, 1.0036] & -- \\
Bottleneck response-task retention &
\(0.9342\pm0.0196\) & [0.9136, 0.9547] & [0.9168, 0.9685] & -- \\
Residual-hidden overall retention &
\(0.5998\pm0.0074\) & [0.5921, 0.6075] & [0.5928, 0.6109] & -- \\
Residual-hidden response-task retention &
\(0.9436\pm0.0322\) & [0.9097, 0.9774] & [0.9047, 0.9958] & -- \\
Linear-readout / trained-readout macrostructure ratio &
\(1.0356\pm0.0056\) & [1.0297, 1.0415] & [1.0249, 1.0395] & -- \\
\bottomrule
\end{tabular}
\endgroup
\end{table*}

\begin{table*}[t]
\centering
\caption{\textbf{Hidden-state geometry, composite-score sensitivity and state-only closure diagnostics.} \textbf{a}, Seed-42 hidden dimensionality and diagnostic six-cluster alignment; representation-space clusters do not redefine the empirical \(K=6\) states. \textbf{b}, Macrostructure scores under the primary, alternative-RMSE and leave-one-domain-out definitions. \(B/F\) is bottleneck/full-hidden retention, and headroom retention references each representation's permutation floor. \textbf{c}, Bottleneck metrics relative to 50 within-user marginal-preserving permutations; improvement is oriented so that positive values are favourable. \textbf{d}, Post hoc state-only closure results on seed-42 confirmation data and six-seed changes relative to the regularisation-matched quadratic. Row TV is the only metric for which lower values are favourable; each floor comparison uses the corresponding closure's own permutations. The matched-origin gauge conditions empirical and model updates on the same predicted current state. \textbf{e}, Shared-sample geometry of predictive-state, pure SSL and task-only representations. Canonical correlations and leading-principal-component alignment are six-seed confirmation means \(\pm\) sample standard deviations; nonlinear gains compare feature-matched histogram gradient boosting and ridge probes on 64-component PCA scores. All diagnostics are descriptive.}
\label{tab:supp_representation_diagnostics}
\begingroup
\scriptsize
\setlength{\tabcolsep}{3.5pt}
\renewcommand{\arraystretch}{1.08}

\textbf{a. Dimensionality and diagnostic mesostate alignment}\par\vspace{2pt}

\begin{tabular}{@{}p{0.48\textwidth}ccc@{}}
\toprule
Diagnostic & Training & Validation & Confirmation \\
\midrule
Participation ratio & 11.23 & -- & -- \\
Effective rank & 18.84 & -- & -- \\
TwoNN intrinsic dimension & 8.913 & 7.766 & 7.471 \\
Full hidden NMI / ARI & -- & 0.1878 / 0.0979 & 0.1852 / 0.0950 \\
Bottleneck NMI / ARI & -- & 0.4030 / 0.2446 & 0.4030 / 0.2378 \\
Residual hidden NMI / ARI & -- & 0.0502 / 0.0418 & 0.0482 / 0.0403 \\
\bottomrule
\end{tabular}

\vspace{5pt}
\textbf{b. Composite-score definitions}\par\vspace{2pt}

\resizebox{\textwidth}{!}{%
\begin{tabular}{lrrrrr}
\toprule
Scoring definition & Full hidden & Bottleneck & Residual hidden &
Raw \(B/F\) & Headroom retention \\
\midrule
Primary, RMSE scale 0.15 & 0.8989 & 0.8166 & 0.5370 & 0.9084 & 0.6470 \\
RMSE scale 0.10 & 0.8797 & 0.7931 & 0.5147 & 0.9016 & 0.6348 \\
RMSE scale 0.20 & 0.9105 & 0.8322 & 0.5541 & 0.9140 & 0.6562 \\
Omit coordinate domain & 0.8742 & 0.7659 & 0.5347 & 0.8761 & 0.5737 \\
Omit closure domain & 0.9438 & 0.8852 & 0.5887 & 0.9380 & 0.7173 \\
Omit drift domain & 0.8936 & 0.8417 & 0.4713 & 0.9418 & 0.7647 \\
Omit transition domain & 0.8841 & 0.7734 & 0.5530 & 0.8749 & 0.5358 \\
\bottomrule
\end{tabular}%
}

\vspace{5pt}
\textbf{c. Bottleneck permutation-floor diagnostics}\par\vspace{2pt}

\resizebox{\textwidth}{!}{%
\begin{tabular}{lrrrl}
\toprule
Metric & Observed & Floor median [5--95\%] &
Oriented improvement & Favourable relative to floor \\
\midrule
Coordinate correlation, \(M\) & 0.8837 & 0.6643 [0.6637, 0.6647] & 0.2194 & Yes \\
Coordinate correlation, \(\Psi\) & 0.9905 & 0.6102 [0.6097, 0.6107] & 0.3803 & Yes \\
One-step RMSE, \(M\) & 0.1524 & 0.2368 [0.2366, 0.2369] & 0.0844 & Yes \\
One-step RMSE, \(\Psi\) & 0.0569 & 0.1629 [0.1628, 0.1630] & 0.1060 & Yes \\
Learned-plane drift correlation & 0.5645 & 0.7107 [0.7052, 0.7160] & \(-0.1462\) & No \\
Learned-plane local cosine & 0.4006 & 0.2360 [0.2304, 0.2413] & 0.1646 & Yes \\
Transition mean row TV & 0.1378 & 0.3416 [0.3409, 0.3425] & 0.2038 & Yes \\
Self-transition correlation & 0.8428 & 0.8580 [0.8565, 0.8596] & \(-0.0151\) & No \\
Diagonal-dominance agreement & 1.0000 & 0.6667 [0.6667, 0.6667] & 0.3333 & Yes \\
Top-edge overlap & 1.0000 & 0.6667 [0.6667, 0.6667] & 0.3333 & Yes \\
\bottomrule
\end{tabular}%
}
\vspace{5pt}
\textbf{d. Post hoc state-only closure functional-form and gauge analysis}
\par\vspace{2pt}

\resizebox{\textwidth}{!}{%
\begin{tabular}{lrrrrl}
\toprule
Diagnostic &
Seed-42 quadratic &
Seed-42 alternative &
\(\Delta\) alternative--quadratic [95\% \(t\)] &
Favourable seeds &
Favourable relative to own floor \\
\midrule
Spline: primary-gauge drift \(r\) &
0.5649 & 0.5712 &
\(+0.0199\;[-0.0362,\,0.0761]\) &
5/6 & No \\
Spline: matched-origin drift \(r\) &
0.3415 & 0.5514 &
\(+0.2219\;[0.1542,\,0.2897]\) &
6/6 & No \\
Spline: primary-gauge local cosine &
0.4006 & 0.2406 &
\(-0.1108\;[-0.2133,\,-0.0084]\) &
0/6 & No \\
Spline: primary-gauge self-transition \(r\) &
0.8430 & 0.8545 &
\(-0.0736\;[-0.1493,\,0.0022]\) &
1/6 & Yes \\
Gaussian: primary-gauge self-transition \(r\) &
0.8430 & 0.6692 &
\(-0.2479\;[-0.3491,\,-0.1468]\) &
0/6 & No \\
Gaussian: matched-origin row TV &
0.1699 & 0.0508 &
\(-0.0902\;[-0.1282,\,-0.0523]\) &
6/6 & Yes \\
Gaussian: matched-origin self-transition \(r\) &
0.4958 & 0.9931 &
\(+0.4628\;[0.3754,\,0.5503]\) &
6/6 & Yes \\
\bottomrule
\end{tabular}%
}
\vspace{5pt}
\textbf{e. Hidden-state geometry in objective-control representations}
\par\vspace{2pt}

\resizebox{\textwidth}{!}{%
\begin{tabular}{lrrrrr}
\toprule
Representation &
Canonical \(r_1\) &
Canonical \(r_2\) &
Leading-PC \(|r_M|\) &
Leading-PC \(|r_\Psi|\) &
Seed-42 nonlinear \(\Delta r_M/\Delta r_\Psi\) \\
\midrule
Predictive-state Event-SSL &
\(0.9913\pm0.0004\) &
\(0.9000\pm0.0009\) &
\(0.7890\pm0.0103\) &
\(0.6178\pm0.0239\) &
\(0.0153/0.0041\) \\
Pure SSL &
\(0.9298\pm0.0039\) &
\(0.8165\pm0.0048\) &
\(0.4139\pm0.0493\) &
\(0.5867\pm0.0287\) &
\(0.0405/0.0268\) \\
Task-only &
\(0.8697\pm0.0071\) &
\(0.8102\pm0.0046\) &
\(0.4532\pm0.0383\) &
\(0.2907\pm0.0209\) &
\(0.0622/0.0727\) \\
\bottomrule
\end{tabular}%
}
\endgroup
\end{table*}

\begin{figure*}[t]
    \centering
    \includegraphics[width=\textwidth]{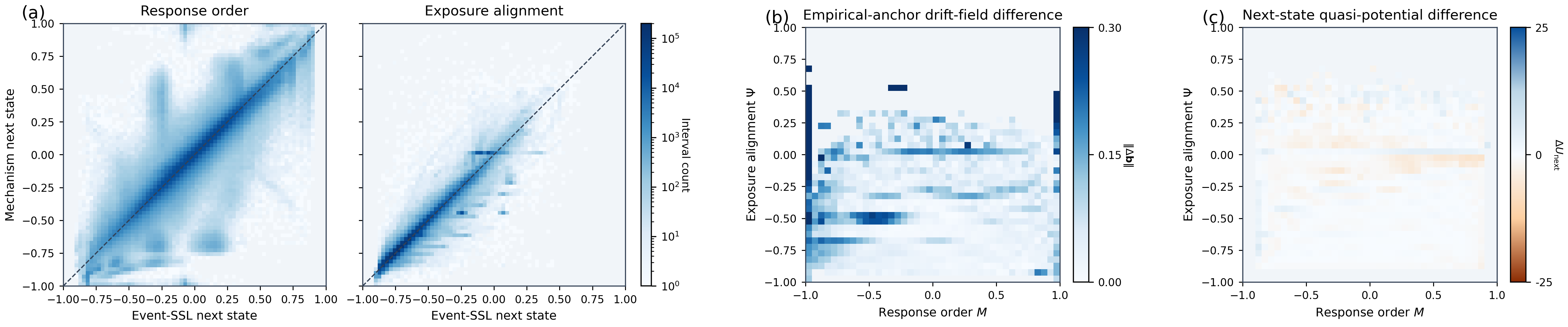}
    \caption{\textbf{Spatial decomposition of agreement between frozen neural
    and mechanistic closures.}
    \textbf{a}, Joint densities of Event-SSL-implied next states (horizontal
    axes) and mechanism-implied next states (vertical axes) for response order
    \(M\) and exposure alignment \(\Psi\) on the common confirmation
    intervals. Dashed lines show identity and colour denotes interval count on
    a logarithmic scale.
    \textbf{b}, Magnitude of the difference between their user-balanced
    empirical-anchor drift fields,
    \(\lVert\mathbf b_{\mathrm{SSL}}-\mathbf b_{\mathrm{mech}}\rVert_2\),
    on jointly supported cells.
    \textbf{c}, Event-SSL-minus-mechanism next-state occupancy-derived
    quasi-potential difference,
    \(\Delta U_{\mathrm{next}}=U_{\mathrm{SSL}}-U_{\mathrm{mech}}\), on common
    next-state support. All panels use frozen seed-42 outputs on \(3{,}233{,}208\) matched intervals from \(56{,}195\) held-out confirmation users; neither model was retrained or fitted to
    the other. Display ranges in \textbf{b} and \textbf{c} are capped at
    \(0.30\) and \(\pm25\), respectively.}
    \label{fig:supp_cross_model_spatial}
\end{figure*}

\subsection*{Supplementary Note 5. Null-referenced downstream recovery and cross-model spatial structure}

The exact cyclic-shift expectation \(\mathbf b_0\) served only as a deterministic decomposition reference; prespecified construction-null inference remained permutation-based. Empirical and model fields were aggregated using identical anchors, weights, grid cells and support. Null subtraction used empirical-anchor Event-SSL fields, leaving learned-plane dynamics as a separate self-consistency result. This post hoc analysis evaluated six fixed models; no model, support definition or evaluation rule was changed.

On confirmation, the mechanism reduced the occupancy-weighted distance to the empirical field from \(0.0907\) for the construction-null expectation to \(0.0217\), achieving null-relative skill \(0.9430\). Event-SSL showed reproducible, axis-specific recovery across six seeds: excess-vector correlation averaged \(0.5639\), weighted local cosine \(0.4077\), amplitude slope \(0.4947\) and \(\Psi\)-specific skill \(0.4660\), with all four quantities positive in every seed. Positive \(\Psi\)-specific skill alongside negative overall and \(M\)-specific skill localised the calibration gap to response order. Event-SSL therefore recovered a reproducible directional and partial-amplitude component of the excess field, a representation-level result distinct from calibrated full-field closure (Supplementary Table~\ref{tab:supp_null_referenced_models}).

Across all \(3{,}233{,}208\) matched confirmation intervals, the independently frozen models agreed more strongly after coarse-graining: next-state correlations were \(0.9238\) and \(0.9779\), population-field vector and speed correlations \(0.8661\) and \(0.8552\), weighted local cosine \(0.7851\), and self-transition correlation \(0.9660\). A separate six-seed common-target analysis retained a smaller positive partial correlation (\(0.2048\)) and positive direct empirical-error correlation, but local residual direction was unstable. Neural--mechanistic agreement is therefore concentrated in leading population structure rather than shared cellwise corrections. The four-term model provides a compact surrogate for Event-SSL's leading macroscopic dynamics, while residual direction and transition routes remain model-specific (Supplementary Fig.~\ref{fig:supp_cross_model_spatial}).

\begin{table*}[t]
\centering
\caption{\textbf{Construction-null-referenced recovery of frozen downstream models.} \textbf{a}, Field recovery from identical empirical anchors. \(D_0=D_w(\mathbf b_0,\mathbf b)\), \(D_k=D_w(\mathbf b_k,\mathbf b)\), \(\Delta D^2=D_k^2-D_0^2\), \(S_k=1-D_k^2/D_0^2\) and \(R_k=D_k/D_0\). Brackets are paired confirmation-user multiplier intervals with frozen matching groups and row-level donor expectations. \textbf{b}, Comparison with a non-negative, no-intercept coordinatewise scaffold rescaling fitted on validation and frozen for confirmation. \textbf{c}, Confirmation geometry under the unchanged training-defined core and shell. \textbf{d}, Six-seed Event-SSL null-referenced calibration; brackets are 95\% Student-\(t\) intervals across seeds. Negative \(\Delta D^2\) and positive skill favour the corresponding model over the construction null. \textbf{e}, Six-seed cross-model alignment after accounting for the common empirical target. The linear benchmark and partial correlation are descriptive, not cell-level inferential nulls, and do not replace the permutation tests or the primary comparison on the full matched confirmation cohort.}
\label{tab:supp_null_referenced_models}
\begingroup
\scriptsize
\setlength{\tabcolsep}{3.1pt}
\renewcommand{\arraystretch}{1.08}

\textbf{a. Primary null-referenced field recovery}\par\vspace{2pt}

\resizebox{\textwidth}{!}{%
\begin{tabular}{llrrrrrrrr}
\toprule
Split & Model & Cells & \(D_0\) & \(D_k\) &
\(\Delta D^2\) [95\%] & \(S_k\) & \(R_k\) & \(S_M\) & \(S_\Psi\) \\
\midrule
validation & Four-term mechanism &
940 & 0.0910 & 0.0213 & \(-0.0078\) & 0.9449 & 0.2346 & 0.7395 & 0.9886 \\
validation & Event-SSL, seed 42 &
940 & 0.0910 & 0.2156 & 0.0382 & \(-4.6216\) & 2.3710 & \(-28.2696\) & 0.4017 \\
confirmation & Four-term mechanism &
954 & 0.0907 & 0.0217 & \(-0.0078\) \([-0.0079, -0.0076]\) &
0.9430 & 0.2388 & 0.7214 & 0.9886 \\
confirmation & Event-SSL, seed 42 &
954 & 0.0907 & 0.2149 & 0.0379 [0.0373, 0.0385] &
\(-4.6108\) & 2.3687 & \(-28.9358\) & 0.4024 \\
\bottomrule
\end{tabular}%
}

\vspace{5pt}
\textbf{b. Specificity beyond the rescaled scaffold}\par\vspace{2pt}

\resizebox{\textwidth}{!}{%
\begin{tabular}{lrrrrr}
\toprule
Model & \(\Delta D^2_{\mathrm{diag}}\) [95\%] &
\(r_{\mathrm{exc}}\) & \(r_{\lVert\mathbf b\rVert}\) &
\(c_{\mathrm{local}}\) & \(\beta_{\mathrm{exc}}\) \\
\midrule
Four-term mechanism &
\(-0.0070\) \([-0.0072, -0.0069]\) & 0.8361 & 0.8679 & 0.9473 & 0.9748 \\
Event-SSL, seed 42 &
0.0387 [0.0380, 0.0392] & 0.5531 & 0.6132 & 0.3427 & 0.5227 \\
\bottomrule
\end{tabular}%
}

\vspace{5pt}
\textbf{c. Frozen-support and core--shell geometry}\par\vspace{2pt}

\resizebox{\textwidth}{!}{%
\begin{tabular}{lrrrrr}
\toprule
Field & \(f_{\nabla\cdot\mathbf b<0}\) &
\(\langle\nabla\cdot\mathbf b\rangle_w\) &
\(f_{\mathrm{in}}\) & \(c_{\mathrm{in}}\) &
\(R_{\mathrm{core/shell}}\) \\
\midrule
Empirical excess field & 0.6858 & \(-0.0613\) & 0.7753 & 0.3555 & 0.7303 \\
Mechanism correction & 0.7782 & \(-0.0770\) & 0.7584 & 0.4086 & 0.7248 \\
Event-SSL correction & 0.7764 & \(-0.2040\) & 0.7176 & 0.2862 & 0.4470 \\
\bottomrule
\end{tabular}%
}
\vspace{5pt}
\textbf{d. Six-seed Event-SSL null-referenced calibration}\par\vspace{2pt}

\resizebox{\textwidth}{!}{%
\begin{tabular}{lrl}
\toprule
Metric & Mean [95\% seed-\(t\)] & Seed consistency \\
\midrule
\(\Delta D^2\) &
\(0.0377\;[0.0358,\,0.0397]\) &
\(6/6>0\) \\

Overall null-relative skill &
\(-4.5869\;[-4.8247,\,-4.3490]\) &
\(6/6<0\) \\

\(M\)-specific null-relative skill &
\(-29.1046\;[-30.1357,\,-28.0734]\) &
\(6/6<0\) \\

\(\Psi\)-specific null-relative skill &
\(0.4660\;[0.3802,\,0.5519]\) &
\(6/6>0\) \\

Excess-vector correlation &
\(0.5639\;[0.5507,\,0.5770]\) &
\(6/6>0\) \\

Excess weighted local cosine &
\(0.4077\;[0.3528,\,0.4625]\) &
\(6/6>0\) \\

Excess amplitude slope &
\(0.4947\;[0.4323,\,0.5571]\) &
\(6/6>0\) \\
\bottomrule
\end{tabular}%
}
\vspace{5pt}
\textbf{e. Six-seed common-target-conditioned cross-model agreement}\par\vspace{2pt}

\resizebox{\textwidth}{!}{%
\begin{tabular}{lr}
\toprule
Metric & Mean [95\% seed-\(t\)] \\
\midrule
Raw mechanism--Event-SSL field \(r\) &
\(0.8762\;[0.8694,\,0.8830]\) \\

Descriptive linear common-target benchmark &
\(0.8481\;[0.8414,\,0.8548]\) \\

Raw minus common-target benchmark &
\(0.0281\;[0.0265,\,0.0297]\) \\

Partial \(r_{\mathrm{mech,SSL}\mid\mathrm{emp}}\) &
\(0.2048\;[0.1920,\,0.2177]\) \\

Direct empirical-error vector \(r\) &
\(0.2140\;[0.2026,\,0.2253]\) \\

Direct empirical-error local cosine &
\(0.0618\;[-0.0233,\,0.1468]\) \\
\bottomrule
\end{tabular}%
}
\endgroup
\end{table*}

\subsection*{Supplementary Note 6. Learner-level uncertainty, estimands and field-grid robustness}

Positive-exponential learner multipliers varied learner composition under the prespecified analysis. Their 2.5--97.5\% ranges are sensitivity distributions, distinct from random-seed, mechanism-selection and construction-null uncertainty; the unperturbed estimate need not lie at their centre. Empirical contraction, fixed-model fields, control directions and cross-model agreement retained their stated directions (Supplementary Table~\ref{tab:supp_global_robustness}a). Learner-cluster residence uncertainty is reported separately in Supplementary Table~\ref{tab:supp_kinetic_robustness}.

A separate paired learner bootstrap quantified confirmation-sample uncertainty for frozen field and transition summaries. Fixed-support drift-vector intervals were \(0.9351\)--\(0.9546\) for mechanism recovery, \(0.6653\)--\(0.7089\) for Event-SSL learned-plane recovery and \(0.8569\)--\(0.8747\) for cross-model field agreement. Reapplying the count threshold preserved all field directions and retained a median \(99.95\%\) of primary-support occupancy mass (median support Jaccard, \(0.961\)). Across the six fixed states, the cross-model persistence profile had Pearson \(r=0.9660\), Spearman \(\rho=1.000\) and leave-one-state-out Pearson correlations of \(0.9608\)--\(0.9841\). These statewise summaries remain descriptive; their learner intervals condition on the fixed states (Supplementary Table~\ref{tab:supp_headline_uncertainty}).

Strict user-equal estimands exposed a specific weighting boundary. Response-order readout was weaker when every learner contributed one total unit, whereas exposure-alignment readout remained nearly unchanged. Per-learner row-normalised transitions retained operational diagonal dominance in five of six empirical states. The exception was \(S_0\), whose self-transition probability decreased from \(0.3623\) under interval counts to \(0.0860\) under strict user-equal weighting. Model and cross-model fields already aggregated with equal total learner mass were unchanged by construction. Under the strict user-equal transition estimand, the cross-model self-transition correlation remained \(0.9489\).

Across \(30\times30\), \(40\times40\), \(50\times50\) and
\(40\times40\) interior-only evaluations, empirical replication, mechanism
recovery, Event-SSL learned-plane recovery, cross-model agreement and the
principal hidden-representation fields retained their directional
conclusions (Supplementary Table~\ref{tab:supp_global_robustness}b). Grid
variation changed local resolution but did not reverse the reported field
relationships.

\begin{table*}[t]
\centering
\caption{\textbf{Learner-composition, strict user-equal and field-grid sensitivity.}
\textbf{a}, Primary values, 2.5--97.5\% ranges from 1,000
positive-exponential learner multipliers and strict user-equal alternatives. The multiplier ranges are learner-composition sensitivity distributions, not confidence intervals centred on the unperturbed estimate; the primary value therefore need not lie within the reported range. ``Same'' denotes fields already user-balanced by construction. The strict transition estimand averages each learner's row-normalised transition row over learners visiting the origin state.
\textbf{b}, Ranges across \(30\times30\), \(40\times40\),
\(50\times50\) and \(40\times40\) interior-only evaluations under unchanged
count thresholds. \(c_w\) is the occupancy-weighted local cosine.}
\label{tab:supp_global_robustness}
\begingroup
\scriptsize
\setlength{\tabcolsep}{3.6pt}
\renewcommand{\arraystretch}{1.08}

\textbf{a. Learner composition and strict user-equal estimands}\par\vspace{2pt}

\resizebox{\textwidth}{!}{%
\begin{tabular}{lccc}
\toprule
Quantity & Primary & Learner-multiplier range & Strict user-equal \\
\midrule
\multicolumn{4}{l}{\textit{Empirical effective dynamics}} \\
Training--validation mean local drift cosine & 0.9657 & [0.9358, 0.9498] & 0.9625 \\
Validation negative-divergence occupancy & 0.7762 & [0.7556, 0.8016] & 0.7690 \\
Validation frozen-shell inward fraction & 0.6693 & [0.6521, 0.6800] & 0.6676 \\
Validation core-to-shell speed ratio & 0.5226 & [0.5114, 0.5349] & 0.5433 \\
Diagonal-dominant empirical rows & 6/6 & -- & 5/6 \\
\(S_0\) self-transition probability &
0.3623 &
[0.3261, 0.4150] &
0.0860 \\
\addlinespace
\multicolumn{4}{l}{\textit{Event-SSL interval-level readout}} \\
Current-state correlation, \(M\) & 0.8837 & [0.8792, 0.8902] & 0.7046 \\
Current-state correlation, \(\Psi\) & 0.9905 & [0.9897, 0.9911] & 0.9884 \\
One-step RMSE, \(M\) & 0.1100 & [0.1084, 0.1116] & 0.2990 \\
One-step RMSE, \(\Psi\) & 0.0353 & [0.0347, 0.0361] & 0.0935 \\
\addlinespace
\multicolumn{4}{l}{\textit{Frozen fields, controls and cross-model agreement}} \\
Shuffled-to-ordered transfer learned-plane drift correlation & \(-0.4191\) & \([-0.4222, -0.3733]\) & Same \\
Main minus support-alignment-randomised inward fraction & 0.2047 & [0.2012, 0.2306] & Same \\
\bottomrule
\end{tabular}%
}

\vspace{5pt}
\textbf{b. Field-grid sensitivity}\par\vspace{2pt}

\resizebox{\textwidth}{!}{%
\begin{tabular}{lrrr}
\toprule
Comparison or representation & Drift-vector \(r\) &
\(c_w\) & Drift-speed \(r\) \\
\midrule
Training--validation empirical field & 0.9350--0.9639 &
0.9651--0.9839 & 0.9228--0.9527 \\
Mechanism versus empirical field & 0.9375--0.9492 &
0.9317--0.9648 & 0.9369--0.9504 \\
Event-SSL learned plane versus empirical field & 0.6793--0.7604 &
0.8479--0.9014 & 0.4897--0.5749 \\
Mechanism versus Event-SSL empirical-anchor field & 0.8561--0.8661 &
0.7148--0.7851 & 0.8283--0.8698 \\
Full hidden representation & 0.6894--0.7537 &
0.9142--0.9419 & 0.6223--0.6748 \\
Two-coordinate bottleneck & 0.5616--0.5972 &
0.4004--0.5057 & 0.6454--0.6677 \\
\bottomrule
\end{tabular}%
}
\endgroup
\end{table*}

\begin{table*}[t]
\centering
\caption{\textbf{Paired learner-cluster bootstrap uncertainty for frozen field and transition summaries.} \textbf{a}, Field estimates from 2,000 whole-learner bootstrap replicates. Fixed-support intervals are primary; support-reselected intervals reapply the unchanged 30-transition threshold. \textbf{b}, Transition summaries under the fixed empirical \(K=6\) partition. Learner intervals for six-state correlations condition on those fixed states. \textbf{c}, Self-transition probabilities underlying the cross-model persistence comparison.}
\label{tab:supp_headline_uncertainty}
\begingroup
\scriptsize
\setlength{\tabcolsep}{3.2pt}
\renewcommand{\arraystretch}{1.06}

\textbf{a. Frozen field summaries}\par\vspace{2pt}

\resizebox{\textwidth}{!}{%
\begin{tabular}{llrrr}
\toprule
Comparison & Metric & Point &
Fixed-support 95\% CI &
Support-reselected 95\% interval \\
\midrule
Mechanism versus empirical & Drift-vector \(r\) &
0.9457 & [0.9351, 0.9546] & [0.8963, 0.9307] \\
& Drift-speed \(r\) &
0.9423 & [0.9288, 0.9533] & [0.8927, 0.9347] \\
& \(c_w\) &
0.9542 & [0.9452, 0.9632] & [0.9350, 0.9531] \\
\addlinespace
Event-SSL anchor vs empirical & Drift-vector \(r\) &
0.8802 & [0.8699, 0.8898] & [0.8436, 0.8772] \\
& Drift-speed \(r\) &
0.8649 & [0.8452, 0.8823] & [0.8060, 0.8712] \\
& \(c_w\) &
0.8192 & [0.8100, 0.8284] & [0.8066, 0.8253] \\
\addlinespace
Event-SSL learned plane vs empirical & Drift-vector \(r\) &
0.6877 & [0.6653, 0.7089] & [0.6090, 0.6596] \\
& Drift-speed \(r\) &
0.4926 & [0.4513, 0.5318] & [0.4269, 0.5150] \\
& \(c_w\) &
0.8630 & [0.8464, 0.8795] & [0.8257, 0.8786] \\
\addlinespace
Mechanism vs Event-SSL anchor & Drift-vector \(r\) &
0.8661 & [0.8569, 0.8747] & [0.8365, 0.8603] \\
& Drift-speed \(r\) &
0.8552 & [0.8393, 0.8696] & [0.8132, 0.8629] \\
& \(c_w\) &
0.7851 & [0.7774, 0.7928] & [0.7757, 0.7911] \\
\bottomrule
\end{tabular}%
}

\vspace{5pt}
\textbf{b. Transition and persistence summaries}\par\vspace{2pt}

\resizebox{\textwidth}{!}{%
\begin{tabular}{lrrr}
\toprule
Comparison &
Mean row TV [95\% CI] &
Self-transition Pearson \(r\) [95\% learner interval] &
Self-transition Spearman \(\rho\) [95\% learner interval] \\
\midrule
Mechanism versus empirical &
0.1021 [0.0983, 0.1059] &
0.9870 [0.9849, 0.9887] &
1.000 [1.000, 1.000] \\
Event-SSL anchor vs empirical &
0.1512 [0.1464, 0.1561] &
0.9529 [0.9492, 0.9564] &
1.000 [0.9429, 1.000] \\
Event-SSL learned plane vs empirical &
0.09417 [0.09170, 0.09663] &
0.8462 [0.7908, 0.8879] &
0.8857 [0.7143, 0.9429] \\
Mechanism vs Event-SSL anchor &
0.1497 [0.1447, 0.1547] &
0.9660 [0.9613, 0.9701] &
1.000 [0.9429, 1.000] \\
\bottomrule
\end{tabular}%
}

\vspace{5pt}
\textbf{c. Cross-model statewise persistence}\par\vspace{2pt}

\resizebox{0.64\textwidth}{!}{%
\begin{tabular}{lrrr}
\toprule
State &
Mechanism \(P_{ii}\) &
Event-SSL anchor \(P_{ii}\) &
Difference \\
\midrule
\(S_0\) & 0.3620 & 0.3167 & +0.0453 \\
\(S_1\) & 0.9592 & 0.8070 & +0.1522 \\
\(S_2\) & 0.9665 & 0.8833 & +0.0832 \\
\(S_3\) & 0.9947 & 0.9773 & +0.0175 \\
\(S_4\) & 0.8103 & 0.6639 & +0.1465 \\
\(S_5\) & 0.5508 & 0.3402 & +0.2105 \\
\bottomrule
\end{tabular}%
}

\endgroup
\end{table*}


\begin{thebibliography}{99}

\bibitem{Perdomo2020}
Perdomo, J. C., Zrnic, T., Mendler-D{\"u}nner, C. \& Hardt, M. Performative prediction. In \textit{Proc. 37th International Conference on Machine Learning}, Vol. 119, 7599--7609 (PMLR, 2020).

\bibitem{Glickman2025}
Glickman, M. \& Sharot, T. How human--AI feedback loops alter human perceptual, emotional and social judgements. \textit{Nat. Hum. Behav.} \textbf{9}, 345--359 (2025).

\bibitem{Barnett2023}
Barnett, L. \& Seth, A. K. Dynamical independence: discovering emergent macroscopic processes in complex dynamical systems. \textit{Phys. Rev. E} \textbf{108}, 014304 (2023).

\bibitem{Moore2025}
Moore, S. A., Mann, B. P. \& Chen, B. Automated global analysis of experimental dynamics through low-dimensional linear embeddings. \textit{npj Complex} \textbf{2}, 36 (2025).

\bibitem{Chen2024}
Chen, X. et al. Constructing custom thermodynamics using deep learning. \textit{Nat. Comput. Sci.} \textbf{4}, 66--85 (2024).

\bibitem{Fabiani2024}
Fabiani, G. et al. Task-oriented machine learning surrogates for tipping points of agent-based models. \textit{Nat. Commun.} \textbf{15}, 4117 (2024).

\bibitem{Recanatesi2021}
Recanatesi, S. et al. Predictive learning as a network mechanism for extracting low-dimensional latent space representations. \textit{Nat. Commun.} \textbf{12}, 1417 (2021).

\bibitem{Lu2025}
Lu, M., Marghetis, T. \& Yang, V. C. A first-principles mathematical model integrates the disparate timescales of human learning. \textit{npj Complex} \textbf{2}, 15 (2025).

\bibitem{Choi2020}
Choi, Y. et al. EdNet: a large-scale hierarchical dataset in education. In \textit{Artificial Intelligence in Education}, \textit{Lect. Notes Comput. Sci.} \textbf{12164}, 69--73 (Springer, 2020).

\bibitem{Nanda2023}
Nanda, N., Lee, A. \& Wattenberg, M. Emergent linear representations in world models of self-supervised sequence models. In \textit{Proc. 6th BlackboxNLP Workshop}, 16--30 (Association for Computational Linguistics, 2023).

\bibitem{Park2024}
Park, K., Choe, Y. J. \& Veitch, V. The linear representation hypothesis and the geometry of large language models. In \textit{Proc. 41st International Conference on Machine Learning}, Vol. 235, 39643--39666 (PMLR, 2024).

\bibitem{Simon2026}
Simon, J. et al. There will be a scientific theory of deep learning. Preprint at arXiv:2604.21691 (2026).

\bibitem{KaplanMeier1958}
Kaplan, E. L. \& Meier, P. Nonparametric estimation from incomplete observations. \textit{J. Am. Stat. Assoc.} \textbf{53}, 457--481 (1958).

\bibitem{Greenwood1926}
Greenwood, M. \textit{A Report on the Natural Duration of Cancer}. Reports on Public Health and Medical Subjects No. 33, 1--26 (His Majesty's Stationery Office, 1926).

\bibitem{Benjamini1995}
Benjamini, Y. \& Hochberg, Y. Controlling the false discovery rate: a practical and powerful approach to multiple testing. \textit{J. R. Stat. Soc. Ser. B} \textbf{57}, 289--300 (1995).

\bibitem{Lin1991}
Lin, J. Divergence measures based on the Shannon entropy. \textit{IEEE Trans. Inf. Theory} \textbf{37}, 145--151 (1991).

\bibitem{Lucke2024}
L{\"u}cke, M., Winkelmann, S., Heitzig, J., Molkenthin, N. \& Koltai, P. Learning interpretable collective variables for spreading processes on networks. \textit{Phys. Rev. E} \textbf{109}, L022301 (2024).

\bibitem{Mardt2018}
Mardt, A., Pasquali, L., Wu, H. \& No{\'e}, F. VAMPnets for deep learning of molecular kinetics. \textit{Nat. Commun.} \textbf{9}, 5 (2018).

\bibitem{Wang2019}
Wang, Y., Ribeiro, J. M. L. \& Tiwary, P. Past--future information bottleneck for sampling molecular reaction coordinate simultaneously with thermodynamics and kinetics. \textit{Nat. Commun.} \textbf{10}, 3573 (2019).

\bibitem{Gao2024Dynamics}
Gao, T.-T., Barzel, B. \& Yan, G. Learning interpretable dynamics of stochastic complex systems from experimental data. \textit{Nat. Commun.} \textbf{15}, 6029 (2024).

\bibitem{Machta2013}
Machta, B. B., Chachra, R., Transtrum, M. K. \& Sethna, J. P. Parameter space compression underlies emergent theories and predictive models. \textit{Science} \textbf{342}, 604--607 (2013).

\bibitem{Transtrum2014}
Transtrum, M. K. \& Qiu, P. Model reduction by manifold boundaries. \textit{Phys. Rev. Lett.} \textbf{113}, 098701 (2014).

\bibitem{Transtrum2015}
Transtrum, M. K. et al. Perspective: sloppiness and emergent theories in physics, biology, and beyond. \textit{J. Chem. Phys.} \textbf{143}, 010901 (2015).

\bibitem{Kornblith2019}
Kornblith, S., Norouzi, M., Lee, H. \& Hinton, G. Similarity of neural network representations revisited. In \textit{Proc. 36th International Conference on Machine Learning}, Vol. 97, 3519--3529 (PMLR, 2019).

\bibitem{Huh2024}
Huh, M., Cheung, B., Wang, T. \& Isola, P. Position: the Platonic representation hypothesis. In \textit{Proc. 41st International Conference on Machine Learning}, Vol. 235, 20617--20642 (PMLR, 2024).

\bibitem{Beretta2025}
Beretta, A. F., Zanchetta, D., Bontorin, S. \& De Domenico, M. Latent geometry emerging from network-driven processes. \textit{npj Complex} \textbf{2}, 37 (2025).

\bibitem{Bartolozzi2022}
Bartolozzi, C., Indiveri, G. \& Donati, E. Embodied neuromorphic intelligence. \textit{Nat. Commun.} \textbf{13}, 1024 (2022).

\bibitem{Hafner2025}
Hafner, D., Pasukonis, J., Ba, J. \& Lillicrap, T. Mastering diverse control tasks through world models. \textit{Nature} \textbf{640}, 647--653 (2025).

\bibitem{Madduri2026}
Madduri, M. M. et al. Computational framework to predict and shape human--machine interactions in closed-loop, co-adaptive neural interfaces. \textit{Nat. Mach. Intell.} \textbf{8}, 372--387 (2026).

\bibitem{MacQueen1967}
MacQueen, J. Some methods for classification and analysis of multivariate observations. In \textit{Proc. Fifth Berkeley Symposium on Mathematical Statistics and Probability}, Vol. 1, 281--297 (University of California Press, 1967).

\bibitem{Pedregosa2011}
Pedregosa, F. et al. Scikit-learn: machine learning in Python. \textit{J. Mach. Learn. Res.} \textbf{12}, 2825--2830 (2011).

\bibitem{Efron1979}
Efron, B. Bootstrap methods: another look at the jackknife. \textit{Ann. Stat.} \textbf{7}, 1--26 (1979).

\bibitem{Hastie2009}
Hastie, T., Tibshirani, R. \& Friedman, J. \textit{The Elements of Statistical Learning: Data Mining, Inference, and Prediction} 2nd edn (Springer, 2009).

\bibitem{Weinberger2009}
Weinberger, K. Q., Dasgupta, A., Attenberg, J., Langford, J. \& Smola, A. J. Feature hashing for large scale multitask learning. In \textit{Proc. 26th International Conference on Machine Learning}, 1113--1120 (ACM, 2009).

\bibitem{Cho2014}
Cho, K. et al. Learning phrase representations using RNN encoder--decoder for statistical machine translation. In \textit{Proc. 2014 Conference on Empirical Methods in Natural Language Processing}, 1724--1734 (Association for Computational Linguistics, 2014).

\bibitem{Ba2016}
Ba, J. L., Kiros, J. R. \& Hinton, G. E. Layer normalization. Preprint at arXiv:1607.06450 (2016).

\bibitem{Hendrycks2016}
Hendrycks, D. \& Gimpel, K. Gaussian error linear units (GELUs). Preprint at arXiv:1606.08415 (2016).

\bibitem{Srivastava2014}
Srivastava, N., Hinton, G., Krizhevsky, A., Sutskever, I. \& Salakhutdinov, R. Dropout: a simple way to prevent neural networks from overfitting. \textit{J. Mach. Learn. Res.} \textbf{15}, 1929--1958 (2014).

\bibitem{Huber1964}
Huber, P. J. Robust estimation of a location parameter. \textit{Ann. Math. Stat.} \textbf{35}, 73--101 (1964).

\bibitem{Loshchilov2019}
Loshchilov, I. \& Hutter, F. Decoupled weight decay regularization. In \textit{International Conference on Learning Representations} (2019).

\bibitem{Paszke2019}
Paszke, A. et al. PyTorch: an imperative style, high-performance deep learning library. In \textit{Advances in Neural Information Processing Systems}, Vol. 32, 8024--8035 (Curran Associates, 2019).

\bibitem{Hoerl1970}
Hoerl, A. E. \& Kennard, R. W. Ridge regression: biased estimation for nonorthogonal problems. \textit{Technometrics} \textbf{12}, 55--67 (1970).

\bibitem{Sculley2010}
Sculley, D. Web-scale \(k\)-means clustering. In \textit{Proc. 19th International Conference on World Wide Web}, 1177--1178 (ACM, 2010).

\bibitem{Strehl2002}
Strehl, A. \& Ghosh, J. Cluster ensembles---a knowledge reuse framework for combining multiple partitions. \textit{J. Mach. Learn. Res.} \textbf{3}, 583--617 (2002).

\bibitem{HubertArabie1985}
Hubert, L. \& Arabie, P. Comparing partitions. \textit{J. Classif.} \textbf{2}, 193--218 (1985).

\bibitem{Pearson1901}
Pearson, K. LIII. On lines and planes of closest fit to systems of points in space. \textit{Philos. Mag.} \textbf{2}, 559--572 (1901).

\bibitem{Hotelling1936}
Hotelling, H. Relations between two sets of variates. \textit{Biometrika} \textbf{28}, 321--377 (1936).

\bibitem{Friedman2001}
Friedman, J. H. Greedy function approximation: a gradient boosting machine. \textit{Ann. Stat.} \textbf{29}, 1189--1232 (2001).

\bibitem{RoyVetterli2007}
Roy, O. \& Vetterli, M. The effective rank: a measure of effective dimensionality. In \textit{Proc. 15th European Signal Processing Conference}, 606--610 (2007).

\bibitem{Facco2017}
Facco, E., d'Errico, M., Rodriguez, A. \& Laio, A. Estimating the intrinsic dimension of datasets by a minimal neighborhood information. \textit{Sci. Rep.} \textbf{7}, 12140 (2017).

\end{thebibliography}
\end{document}